\documentclass[10pt,journal,compsoc]{IEEEtran}

\usepackage[utf8]{inputenc} 
\usepackage[T1]{fontenc}    
\usepackage{hyperref}       
\usepackage{url}            
\usepackage{booktabs}       
\usepackage{amsfonts}       
\usepackage{nicefrac}       
\usepackage{microtype}      
\usepackage{xcolor}         
\usepackage{algorithm}
\usepackage{algorithmic}
\usepackage{amssymb}
\usepackage{diagbox}
\usepackage{amsmath}
\usepackage{enumitem}
\usepackage{forest}
\usepackage{adjustbox}
\usepackage{xcolor}
\usepackage{threeparttable}
\usepackage{graphicx}
\usepackage{subfigure}
\usepackage{caption}

\ifCLASSOPTIONcompsoc
  \usepackage[nocompress]{cite}
\else
  \usepackage{cite}
\fi

\ifCLASSINFOpdf

\else

\fi

\begin{document}

\title{Modular Machine Learning: \\ An Indispensable Path towards New-Generation Large Language Models}

\author{
Xin~Wang,~\IEEEmembership{Member,~IEEE,}
        Haoyang~Li,
        Haibo~Chen,
        Zeyang~Zhang
        and~Wenwu~Zhu,~\IEEEmembership{Fellow,~IEEE,}
\IEEEcompsocitemizethanks{\IEEEcompsocthanksitem All authors were with the Department
of Computer Science and Technology, BNRist, Tsinghua University, China, 100084.
E-mail: \{xin\_wang, wwzhu\}@tsinghua.edu.cn,
lihy218@gmail.com,
chb24@mails.tsinghua.edu.cn, zhangzey16@tsinghua.org.cn.
Corresponding author: Wenwu Zhu \protect\\}
\thanks{This work was supported in part by National Natural
Science Foundation of China No. 62222209, Beijing National Research Center for Information Science and Technology (BNRist) under Grant No. BNR2023TD03006, Beijing Key Lab of Networked Multimedia, China.}
}

\markboth{Journal of \LaTeX\ Class Files,~Vol.~14, No.~8, September~2025}%
{Shell \MakeLowercase{\textit{et al.}}: Bare Demo of IEEEtran.cls for Computer Society Journals}

\IEEEtitleabstractindextext{%
\begin{abstract}
Large language models (LLMs) have substantially advanced machine learning research, including natural language processing, computer vision, data mining, etc., yet they still exhibit critical limitations in explainability, reliability, adaptability, and extensibility. In this paper, we overview a promising learning paradigm, i.e., Modular Machine Learning (MML), as an essential approach toward new-generation LLMs capable of addressing these issues. We begin by systematically and comprehensively surveying the existing literature on modular machine learning, with a particular focus on modular data representation and modular models. 
Then, we propose a unified MML framework for LLMs, which decomposes the complex structure of LLMs into three interdependent components: modular representation, modular model, and modular reasoning. 
Specifically, the MML paradigm discussed in this article is able to: i) clarify the internal working mechanism of LLMs through the disentanglement of semantic components; ii) allow for flexible and task-adaptive model design; iii) enable an interpretable and logic-driven decision-making process.
We further elaborate a feasible implementation of MML-based LLMs via leveraging advanced techniques such as disentangled representation learning, neural architecture search and neuro-symbolic learning. Last but not least, we critically identify the remaining key challenges, such as the integration of continuous neural and discrete symbolic processes, joint optimization, and computational scalability, present promising future research directions that deserve further exploration. Ultimately, we believe the integration of the MML with LLMs has the potential to bridge the gap between statistical (deep) learning and formal (logical) reasoning, thereby paving the way for robust, adaptable, and trustworthy AI systems across a wide range of real-world applications.
\end{abstract}
\begin{IEEEkeywords}
Large Language Model, Neuro-Symbolic Learning, Disentangled Representation Learning, Neural Architecture Search.
\end{IEEEkeywords}}

\maketitle

\IEEEdisplaynontitleabstractindextext

\IEEEraisesectionheading{\section{Introduction}\label{sec:intro}}
\label{sec:intro}
\IEEEPARstart{T}{he} advent of Large Language Models (LLMs)~\cite{achiam2023gpt}, epitomized by the likes of ChatGPT~\cite{chatgpt}, has undeniably been a watershed moment in the evolution of the AI landscape. These models have astounded the research community and the general public alike with their seemingly superhuman language-processing capabilities. In a plethora of tasks, they have managed to mimic human-level language understanding and generation with remarkable fidelity. From drafting eloquent essays and engaging in fluent conversations to providing detailed summaries of complex texts, LLMs have proven their capabilities across various scenarios.
This remarkable performance has led some researchers to advocate the “bigger is better” principle. The underlying rationale is that as the model size and volume of training data grow exponentially (i.e., the scaling law), so does the breadth and depth of knowledge that the LLM can encapsulate. This enables the model to handle a wide range of language-related challenges with an accuracy and fluency that were previously unimaginable.
\begin{figure*}[thbp]
    \centering
        \includegraphics[width=1\textwidth]{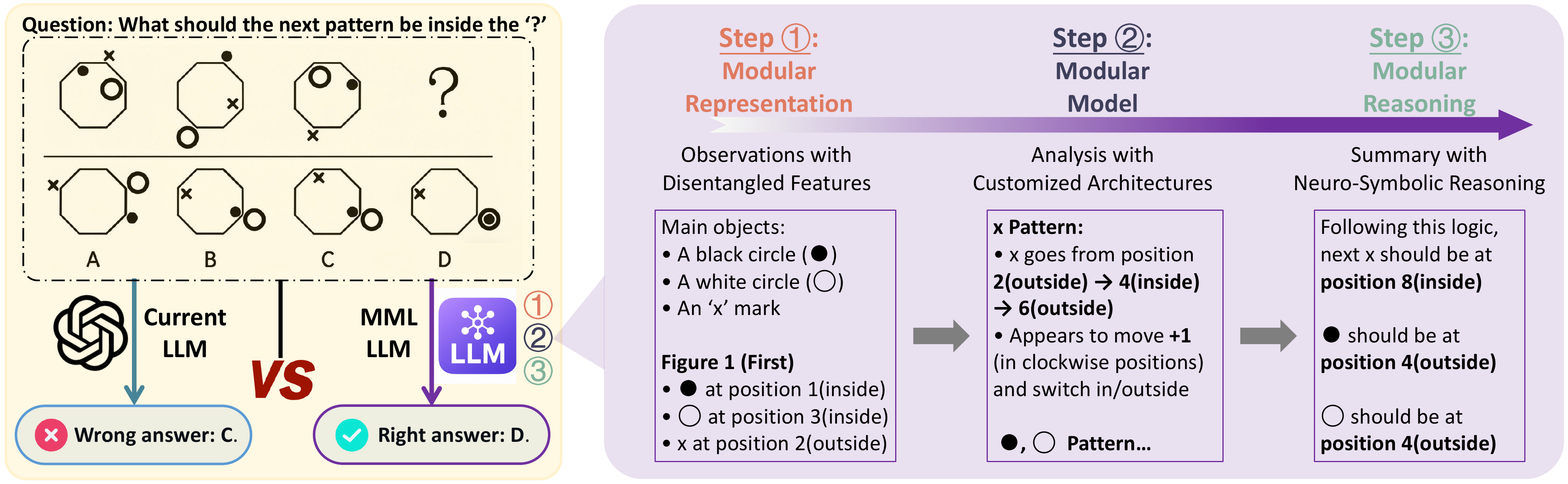}
    \caption{We employ the visual question answering (VQA) task to compare traditional LLMs with our proposed Modular Machine Learning (MML) framework for LLMs. Current LLMs (e.g., ChatGPT-4o) often fail on tasks requiring complex reasoning. MML is able to overcome the weakness by adopting a modular approach as follows, \textbf{Step \textcircled{1}}: Modular Representation first disentangles the visual content to separate and extract key features from image based on the textual question (e.g., identifying the target objects \textbf{X}, \textbf{Solid Circle} and \textbf{Open Circle}); \textbf{Step \textcircled{2}}: Modular Model then utilizes customized neural architectures tailored for different functionalities to analyze sequential position patterns of objects (e.g., tracking the sequential positions of the relevant objects \textbf{X}, \textbf{Solid Circle} and \textbf{Open Circle}); \textbf{Step \textcircled{3}}: Modular Reasoning finally performs logical reasoning to infer the next pattern and summarize the answer (e.g., predict the positions of \textbf{X}, \textbf{Solid Circle} and \textbf{Open Circle} inside the question mark). 
    }
    \label{fig:intro}
    \vspace{-1.5mm}
    \end{figure*}
However, as with any technological marvel, LLMs have their own \textit{Achilles' heels}. A major limitation of LLMs surfaces when it comes to quantitative reasoning tasks. For instance, LLMs often struggle with simple two-digit arithmetic problems. Without human interaction, they have trouble applying basic mathematical principles to obtain the right answers. 
This deficiency highlights the fact that, despite their remarkable linguistic capabilities, their cognitive and logical reasoning abilities, particularly in domains requiring formal and rule-based understanding, remain in need of substantial enhancement.
Consequently, the research community has witnessed a growing trend for augmenting LLMs with additional tools, where the additional tools are designed to fill the gaps in reasoning capabilities. For example, coupling an LLM with a symbolic solver can enhance the LLM's ability to handle mathematical and logical problems. By integrating external knowledge bases, LLMs can access domain-specific information that might not have been part of their original training data as well, thereby enhancing their adaptable problem-solving capabilities. Such hybrid systems could potentially release the full potential of LLMs, allowing them to overcome current limitations and become more versatile and reliable in a wide range of AI applications.

LLMs are expected to be \textit{explainable, reliable, adaptable}, and \textit{extendable} in real-world applications. In many critical domains such as healthcare, finance, and legal systems, the decisions made by LLMs can have far-reaching consequences. For example, when a model provides a diagnosis or financial advice, patients and stakeholders need to understand the underlying rationale. Without explainability, it becomes difficult to validate the accuracy and reliability of the output. Similarly, adaptability is essential as the demands of applications evolve and new tasks emerge; a static LLM would quickly become obsolete. Finally, extensibility is required to keep pace with new scientific discoveries and technological advancements, ensuring that models can incorporate and leverage new knowledge effectively. 

\textbf{Modular Machine Learning (MML)} serves as one indispensable path to achieve the above properties for LLM. By decomposing complex systems into well-defined and interoperable modules, MML directly supports explainability through clearer structure and interpretability, being capable of tracing how specific outputs are produced~\cite{neural2016andreas,kaiser2017one}. It also enhances reliability by allowing modules to be validated and improved independently, reducing the risk of systemic errors. MML fosters adaptability as well, since individual modules can be modified or replaced to meet new requirements without redesigning the entire system~\cite{goyal2022inductive}. Finally, MML is able to promote extensibility, as new scientific methods or domain-specific knowledge can be incorporated as additional modules that integrate seamlessly into the existing architecture~\cite{parascandolo2018learning}. Unlike monolithic end-to-end approaches, the modular perspective provides a structured framework that enhances scalability, transparency, and control—qualities essential for deploying LLMs in real-world, high-stakes environments~\cite{pfeiffer2023modular, khot2020text, oldenburg2024xmodnn}.


In this paper, we present a \textbf{systematic and comprehensive survey of Modular Machine Learning (MML)}, which, to the best of our knowledge, is the first literature review of this paradigm. We begin with a \textit{formal definition} of MML to clarify its conceptual position. We then categorize existing methods into two major groups: (i) \textit{modular data representation}, which aims to separate complex observations into semantically meaningful and independent factors, and (ii) \textit{modular model optimization}, which emphasizes decomposing architectures and reasoning processes into functional modules. Within each group, we review representative approaches under both supervised and unsupervised settings, highlighting their fundamental assumptions, technical designs, and trade-offs. Building upon this taxonomy, we further propose a unified MML framework for LLMs, consisting of three interdependent components: \textit{modular representation}, \textit{modular model}, and \textit{modular reasoning}. For illustrative purposes, we employ the visual question answering (VQA) task as a running example (Fig.~\ref{fig:intro}), which demonstrates how MML disentangles visual content, assembles specialized neural modules, and integrates symbolic reasoning to boost the ability of LLMs in generating interpretable outputs.
To further contextualize the unified MML framework, 
we provide an instantiation of MML under the VQA task with a feasible implementation, where the three components are implemented as follows:
(1) \textit{Disentangled Representation Learning} (DRL) for modular representation enables a more organized and interpretable structure of information within LLMs by separating complex features into independent semantic dimensions, thereby enhancing transparency and controllability. 
(2) \textit{Neural Architecture Search} (NAS) for modular model automates the construction of modularized neural architectures, where different modules are optimized for distinct functions and can be flexibly extended to new tasks. 
(3) \textit{Neuro-Symbolic Learning} (NSL) for modular reasoning integrates symbolic rules with neural inference, thereby formalizing decision-making steps, improving reliability, and facilitating fairness by detecting and correcting potential biases. 
More detailed discussions of implementation aspects will be presented in Section~\ref{sec:mml-implementation}.
In general, this article provides a consolidated view of the field, i) overviewing existing literature comprehensively, and ii) positioning MML as a promising direction for enhancing the \textit{explainability}, \textit{reliability}, \textit{adaptability}, and \textit{extensibility} of large language models and related AI systems.

\textbf{Relations to Relevant Surveys.} Recently, several efforts have focused on topics that are related to a subtopic within a particular component of MML. Concretely, Disentangled Representation Learning (DRL) has been systematically reviewed as a paradigm for separating latent factors into independent and semantically meaningful components~\cite{wang2024disentangled,eddahmani2023unsupervised}, while Neural Architecture Search (NAS) has been extensively surveyed as an automated framework for discovering effective neural architectures under diverse constraints~\cite{elsken2019neural,ren2021comprehensive, baymurzina2022review,white2023neural,chitty2023neural}. 
However, there have been no works that comprehensively discuss key elements such as Neuro-Symbolic Learning (NSL) and Modular Networks etc., not to mention the whole literature on MML.
This paper goes beyond such domain-specific perspectives via providing a unified view that covers the complete collection of works in MML. Moreover, this work examines DRL and NAS from a \textit{Modular Perspective}, which are totally different from those in relevant surveys. 
Last but not least, we propose a unified MML framework for LLMs with feasible implementations in this paper, which has huge potential in reshaping the design principles of next-generation large language models.

In summary, this survey aims to offer a consolidated understanding of the potential and limitations of existing approaches, with the goal of informing the development of more explainable, adaptable, and reliable AI systems, and contributing to the broader pursuit of Artificial General Intelligence (AGI).


\section{Methodology of Modular Machine Learning (MML)}
\label{sec:mml-definition}
We categorize Modular Machine Learning (MML) into two major directions: 
\textbf{Modular Data Representation} and \textbf{Modular Model Optimization}. 
The former emphasizes disentangling complex observations into independent semantic factors at the representation level, 
while the latter emphasizes decomposing architectures and reasoning processes into functional modules. 
Within each category, we distinguish between \textbf{with modular supervision} and \textbf{without modular supervision}, 
depending on whether explicit guidance on the semantics or functionality of modules is available. 
In all cases, the general principle is to optimize the task objective while subject to modularity constraints.

\subsection{Modular Data Representation}

\textbf{Definition (Modular Data Representation).}  
Given input $x \in \mathcal{X}$, a modular data encoder $\mathcal{D}_\theta: \mathcal{X} \to \mathcal{Z}$ maps $x$ into latent factors 
$z=(z_1,\dots,z_n)$. The objective is that each $z_i$ corresponds to a distinct and independent semantic factor $v_i$.

\begin{itemize}
    \item \textbf{With Modular Supervision.}  
    When annotated factors $\{v_i\}$ are available, each latent slot $z_i$ must align with its corresponding semantic factor. 
    The optimization problem is
    \begin{equation}
        \min_{\theta}\; \mathcal{L}_{\text{task}}(z,y)
        \quad \text{s.t. } \; h_i(z_i) = v_i, \;\; \forall i,
    \end{equation}
    where $h_i$ is a prediction head used to enforce the alignment between $z_i$ and its semantic label $v_i$.  
    This ensures that representations are interpretable and directly usable for modular reasoning.

    \item \textbf{Without Modular Supervision.}  
    When no factor labels are available, disentanglement must be induced implicitly. 
    A common constraint is statistical independence across latent dimensions:
    \begin{equation}
        \min_{\theta}\; \mathcal{L}_{\text{task}}(z,y)
        \quad \text{s.t. } \; p(z) = \prod_{i=1}^n p(z_i).
    \end{equation}
    This encourages each latent slot to capture an independent factor of variation, 
    even though the semantics of factors are not explicitly specified.
\end{itemize}

\subsection{Modular Model}

\textbf{Definition (Modular Model).}  
A modular model consists of a set of modules $\{M_k\}_{k=1}^K$ connected under an architecture $\mathcal{A}$. 
Given representation $z$, the overall output is
\begin{equation}
    \hat{y} = \Phi_{\mathcal{A}}\big(M_1(z),\dots,M_K(z)\big),
\end{equation}
where $\Phi_{\mathcal{A}}$ specifies how modules are composed (e.g., routing, stacking, symbolic reasoning).

\begin{itemize}
    \item \textbf{With Modular Supervision.}  
    When the functionality of each module is predefined (e.g., perception, reasoning, or a specific sub-task), the optimization problem becomes
    \begin{equation}
        \min_{\{M_k\},\,\mathcal{A}}\; \mathcal{L}_{\text{task}}(\hat{y},y)
        \quad \text{s.t. } \; M_k \;\text{solves task } t_k, \;\; \forall k,·
    \end{equation}
    where $t_k$ denotes the supervised functionality assigned to module $M_k$. 
    This setting ensures clear interpretability and controllability of the modular architecture.

    \item \textbf{Without Modular Supervision.}  
    When no explicit module-task assignment is provided, both the architecture and module specialization are learned end-to-end. 
    This is typically realized by Neural Architecture Search (NAS) or differentiable routing. The problem is
    \begin{equation}
        \min_{\{M_k\},\,\mathcal{A}}\; \mathcal{L}_{\text{task}}\!\Big(\Phi_{\mathcal{A}}(\{M_k(z)\}),\, y\Big)
        \quad \text{s.t. } \; \mathcal{A} \in \mathcal{S},
    \end{equation}
    where $\mathcal{S}$ is the predefined search space of candidate architectures. 
    Module functionality emerges implicitly through optimization, but interpretability may be weaker compared to the supervised case.
\end{itemize}

\noindent
In summary, both Modular Data Representation and Modular Model Optimization can be realized in supervised or unsupervised manners. 
The supervised versions guarantee interpretability by explicitly aligning slots or modules with semantic factors or tasks, 
while the unsupervised versions emphasize scalability by discovering modular structures automatically through implicit constraints.

\begin{figure*}[thbp]
\centering
\resizebox{0.86\textwidth}{!}{
\input{fig/fig_tree}
}
\caption{Taxonomy of modular machine learning methods.
}
\label{fig-main}
\vspace{-2.5mm}
\end{figure*}

\section{Modular Data Representation}
\label{sec:modular_data}

A central challenge in Modular Machine Learning is to learn modular representations of data, namely to decompose complex observations into independent and semantically meaningful factors. Disentangled Representation Learning (DRL) addresses this challenge by learning representations that separate the underlying generative factors of variation into distinct components. Such disentangled representations enhance explainability, controllability, robustness, and generalization, all of which are desirable for downstream machine learning tasks~\cite{bengio2013representation, higgins2018definition, wang2024drl}. Existing approaches can be broadly divided into those that incorporate explicit priors or supervisory signals to guide the disentanglement process and those that aim to induce modular structure without such specific guidance. In the following subsections, we first review prior-guided approaches, which learn modular data representations under explicit modular supervision, and then turn to prior-agnostic approaches, which aim to discover modular structure without modular supervision.

\begin{table}[ht]
\centering
\caption{Summary of disentangled representation learning methods. 
Abbreviations: 
M = Model (V = VAE-based, G = GAN-based, D = Diffusion-based, P = Pretrained Directions, C = Causal-based, O = Others); 
P = Paradigm (Unsup. = Unsupervised, Sup. = Supervised, WSup. = Weakly Supervised); 
RS = Representation Structure (Dim. = Dimension-wise, Vec. = Vector-wise); 
A = Application (IG = Image Generation, IE = Image Editing, I2I = Image-to-Image Translation, VG = Video Generation, Rec. = Recommendation, Gr. = Graph).}
\vspace{-2mm}
\begin{adjustbox}{max width=0.35\textwidth}
\begin{tabular}{|l|l|l|l|l|}
\hline
\textbf{Method} & \textbf{M} & \textbf{P} & \textbf{RS} & \textbf{A} \\
\hline
$\beta$-TCVAE~\cite{chen2018isolating} & V & Unsup. & Dim. & IG \\
$\beta$-VAE~\cite{higgins2017beta} & V & Unsup. & Dim. & IG \\
FactorVAE~\cite{kim2018disentangling} & V & Unsup. & Dim. & IG \\
JointVAE~\cite{dupont2018jointvae} & V & Unsup. & Dim. & IG \\
RF-VAE~\cite{kim2019relevance} & V & Unsup. & Dim. & IG \\
AnnealedVAE~\cite{burgess2018understanding} & V & Unsup. & Dim. & IG \\
DIP-VAE~\cite{kumar2018variational} & V & Unsup. & Dim. & IG \\
InfoGAN~\cite{chen2016infogan} & G & Unsup. & Dim. & IG \\
VAE-GAN~\cite{larsen2016vaegan} & G & Unsup. & Dim. & IG \\
VON~\cite{zhu2018visual} & G & Unsup. & Dim. & IG \\
StyleSpace~\cite{wu2021stylespace} & G & Unsup. & Dim. & IG \\
Semi-supervised VAE~\cite{locatello2020weakly} & V & Sup. & Dim. & IG \\
CausalVAE~\cite{yang2021causalvae} & C & Unsup. & Dim. & IG \\
DRIT~\cite{lee2018drit} & G & Unsup. & Dim. & I2I \\
Cross-domain I2I~\cite{gonzalez2018crossdomain} & G & Unsup. & Dim. & I2I \\
DRNET~\cite{hsieh2018drnet} & V & Unsup. & Vec. & VG \\
DDPAE~\cite{denton2017ddpae} & V & Unsup. & Vec. & VG \\
MAP-IVR~\cite{tulyakov2018mapivr} & O & Sup. & Vec. & VG \\
MacridVAE~\cite{wang2019macridvae} & V & Sup. & Vec. & Rec. \\
Sequential DRL Rec.~\cite{sun2020seqrec} & V & Sup. & Vec. & Rec. \\
CDR~\cite{ma2020cdr} & V & Sup. & Vec. & Rec. \\
Content-Collaborative DRL~\cite{chen2020ccdr} & O & Unsup. & Vec. & Rec. \\
DGCF~\cite{wang2020dgcf} & O & Unsup. & Vec. & Rec. \\
Unsupervised News Rec.~\cite{li2021newsrec} & O & Unsup. & Vec. & Rec. \\
Node-edge Graph Gen.~\cite{li2019nodeedge} & V & Sup. & Vec. & Gr. \\
Disentangled Graph CL~\cite{li2020graphcl} & V & Sup. & Vec. & Gr. \\
DisenGCN~\cite{ma2019disengcn} & O & Unsup. & Vec. & Gr. \\
OOD-GNN~\cite{li2022oodgnn} & O & Unsup. & Vec. & Gr. \\
\hline
\end{tabular}
\vspace{-2mm}
\end{adjustbox}
\end{table}

\subsection{Modular Data Representation with Modular Supervision}
\label{sec:modular_data_with}
When ground-truth factors of variation are known or labeled for each sample, one can directly supervise the representation to be modular. In the simplest case, each latent dimension $z_i$ is trained to predict or reconstruct its corresponding factor $v_i$, effectively disentangling the representation by design. This fully supervised approach is often feasible on synthetic or controlled datasets where factor labels (e.g., object attributes, orientations, etc.) are provided. For example, on thedSprites benchmark~\cite{dsprites2017} or 3D Shapes dataset~\cite{3dshapes2018} with known generative factors, one can attach a small prediction head $h_i$ to each latent and train $\mathcal{D}_\theta$ such that $z_i$ correlates with the $i$-th true factor (color, shape, size, etc.), yielding an interpretable factorized embedding. Such direct modular supervision guarantees that each latent encodes a distinct semantic concept by construction.

In many cases, explicit factor annotation may be unavailable, but researchers can still leverage partial supervision or structural knowledge of factors to guide disentanglement. One common strategy is to design the model architecture or training procedure to separate known factors. For instance, when certain factor subspaces are conceptually known (e.g., “content” vs. “pose” in images or video), the encoder can be split into sub-networks, and training involves mixing or swapping these parts to enforce factor separation. Peng \textit{et al.}~\cite{peng2017reconstruction} and Denton \& Birodkar~\cite{denton2017unsupervised} both follow this idea: they assume predefined factor subsets and train generative models where swapping latent sub-codes between two inputs yields outputs with exchanged factor values. In Peng \textit{et al.}’s pose-invariant face recognition model, the latent code is partitioned into “identity” and “pose” components; by swapping the pose code between two face images, the decoder generates a new image with one person’s identity and the other’s pose, and reconstruction loss ensures that each partition truly captures the intended factor. Denton \& Birodkar similarly separate content (static appearance) and motion factors in video sequences by enforcing that a “content” latent remains constant across frames while a separate “pose/motion” latent changes over time – thereby disentangling the two without direct labels.

Another strategy uses attribute labels or weak tags on the data to supervise the latent factors. For example, Wang \textit{et al.}~\cite{wang2017tagdisentangled} introduce a Tag‐Disentangled GAN in which images come with attribute tags (such as object identity, viewpoint, illumination). Their framework uses an encoder (disentangling network) and a tag prediction network to ensure each latent component corresponds to a specific tag; a consistency loss between the latent code and the image’s tags drives the encoder to isolate the factors. Kulkarni \textit{et al.}~\cite{kulkarni2015deep} train a convolutional network on images with controlled transformations (like varying pose or lighting in a 3D rendered scene) and enforce that the latent units respond only to the designated transformation that changed across the input batch. By feeding the network pairs or mini-batches of images where only one factor differs, they effectively provide supervisory signals that each latent should capture that factor of variation. Worrall \textit{et al.}~\cite{worrall2017interpretable} also incorporate prior knowledge about transformations (such as rotations) by building an encoder-decoder that includes a continuous latent variable explicitly representing that transformation. Their model, for example, can learn a latent variable for rotation that, when changed, rotates the reconstructed object while other latent variables preserve object identity – demonstrating disentanglement through a form of supervised equivariance.

Modular data representation with supervision leverages any available factor-specific information – be it direct labels, known pairing of data differing in one factor, or architectural partitioning of latent spaces – to impose a one-to-one correspondence between latent dimensions and semantic factors. This yields highly interpretable representations and has been shown to facilitate tasks like zero-shot reasoning and factor-aware image editing, since each $z_i$ can be manipulated independently, knowing its semantic meaning.

\subsection{Modular Data Representation without Modular Supervision}
\label{sec:modular_data_without}
When factor annotations are not available, the model must discover a factorized representation on its own. Unsupervised disentanglement is typically achieved by adding constraints or penalties that bias the encoder $\mathcal{D}_\theta$ to distribute information across $z_i$ in an independent manner. A widely used principle is to encourage the latent distribution $p(z)$ to factorize into $\prod_i p(z_i)$. In practice, methods enforce this by penalizing dependencies among $z_i$ in the training objective. One notable example is the $\beta$-VAE of Higgins \textit{et al.}~\cite{higgins2017beta}, which modifies the Variational Autoencoder loss by up-weighting the KL-divergence term that pushes the posterior towards a factorized normal prior. By increasing this weight ($\beta$), the $\beta$-VAE sacrifices some reconstruction fidelity in exchange for more independent latent factors, often succeeding in uncovering separate generative factors without any labels. 

Following this idea, a number of VAE-based approaches introduced more targeted independence penalties. DIP-VAE (Disentangled Inferred Prior) by Kumar \textit{et al.}~\cite{kumar2018variational} adds a regularizer that explicitly matches the aggregated posterior $q(z)$ to a factorized prior by minimizing off-diagonal covariance in $q(z)$ – essentially decorrelating the latent dimensions. FactorVAE by Kim and Mnih~\cite{kim2018disentangling} and the closely related $\beta$-TCVAE by Chen \textit{et al.}~\cite{chen2018isolating} decompose the VAE’s objective to isolate the total correlation term (which measures latent dependence) and directly penalize it. FactorVAE uses an adversarial classifier to estimate and minimize the total correlation among latent dimensions.
Building on this, $\beta$-TCVAE~\cite{chen2018isolating} decomposes the KL divergence into mutual information, total correlation, and dimension-wise divergence, applying distinct weights to explicitly balance information preservation and disentanglement. These approaches share a unifying modularity perspective: they explicitly separate different roles of the latent dimensions—some responsible for semantic independence, others for information retention. Beyond continuous factors, JointVAE~\cite{dupont2018jointvae} extends this modularization principle by disentangling both continuous and discrete generative factors, showing that modular latent variables can flexibly adapt to heterogeneous semantics. RF-VAE~\cite{kim2019relevance} introduces relevance indicators that filter out nuisance dimensions, effectively pruning the representation space into a cleaner modular structure. Such mechanisms resonate with the MML view, where representations are not only factorized but also selectively routed into relevant modules for downstream reasoning.

Recent works further incorporate group-theoretic principles into VAE-based DRL, enforcing equivariance between latent transformations and symmetry groups of the data. From the modularity perspective, this offers a principled way to align latent modules with structured semantic transformations, enabling systematic generalization and compositional reasoning. Likewise, extensions of VAEs to sequential data disentangle time-varying and time-invariant modules~\cite{li2018disentangled}, showing how modular latent variables can be dynamically partitioned for tasks such as video or speech modeling.
Meanwhile, InfoGAN by Chen \textit{et al.}~\cite{chen2016infogan} augments a Generative Adversarial Network with an information maximization objective: it defines a subset of latent variables $z_i$ as ``interpretable'' factors and adds a mutual information term to ensure these variables can be inferred from the generated data. By maximizing $I(z_i; x)$ for each factor while using a factorized prior on $z$, InfoGAN encourages each $z_i$ to correspond to a distinct salient feature in the data (such as digit rotation or stroke thickness in handwritten digits), again without any supervision on those features.

While these purely unsupervised methods can disentangle factors in idealized settings, in general the problem is ill-posed: there is no guarantee that independent components of $z$ will align with meaningful real-world factors without additional inductive biases. A theoretical study by Locatello \textit{et al.}~\cite{locatello2019challenging} demonstrated that unsupervised disentanglement is fundamentally impossible up to permutation/rotation of latent space, unless the model or data distribution incorporates appropriate constraints. This insight has driven researchers to explore weak supervision and richer assumptions in otherwise unsupervised settings. For instance, some approaches use grouped observations: if we know two inputs share the same value of certain latent factors (e.g., two images of the same object with different backgrounds), the model can be trained to encode them with identical values in the corresponding latent slots, thereby disentangling that factor without explicit labels. Such weakly supervised strategies include learning from temporal sequences or image pairs where some factors remain constant and only others vary. By leveraging temporal coherence or Siamese-like comparisons, the encoder is guided to isolate the invariant content from the changing factors. Recent works build on this idea: Locatello \textit{et al.}~\cite{locatello2020weak} and others propose methods that use pairwise similarity judgments or group information to achieve disentanglement without direct factor annotations, bridging the gap between fully unsupervised and fully supervised approaches.

Diffusion models, which progressively denoise data from Gaussian noise to reconstruct high-fidelity samples, have emerged as a promising framework for disentangled representation learning. Compared to VAEs and GANs, diffusion models offer enhanced generative stability and yield richer intermediate representations, creating novel inductive biases that naturally support modularization of both representation and reasoning. Under the Modular Machine Learning (MML) paradigm, \emph{diffusion-based DRL} focuses on discovering modular latent directions embedded within the denoising trajectory—enabling controllable and interpretable transformations without compromising generation quality.

A representative work is DisDiff~\cite{wang2023disdiff}, which introduces disentanglement into diffusion by enforcing independence among latent embeddings used in score-based generation. DisDiff regularizes the training objective so that distinct latent factors correspond to semantically independent modifications of the generated sample (e.g., color, shape, or texture). From the MML perspective, this establishes disentangled modules within the diffusion trajectory: each module exerts an independent, compositional influence on the generative process. Such modular control enhances counterfactual reasoning, as one can intervene on a single latent module to obtain causal transformations in the output. Another example is DisenBooth~\cite{gao2023disenbooth}, which adapts diffusion-based fine-tuning for text-to-image generation. By constraining optimization to disentangled subspaces, DisenBooth ensures that personalized fine-tuning (e.g., learning a new object or style) does not interfere with unrelated semantic factors. This separation is crucial for modular learning: disentangled modules act as isolated slots, allowing new knowledge to be integrated locally without disrupting existing factors. In practice, this modularity prevents catastrophic entanglement and supports compositional reasoning, such as combining newly learned styles with pre-existing object semantics. Extensions to video generation, such as VideoDreamer~\cite{zhang2023videodreamer}, further highlight the modular perspective. By disentangling temporal and spatial modules within the diffusion process, these methods enable independent control of motion and appearance. This aligns with the MML view of modular reasoning, where different modules are responsible for orthogonal factors and can be recombined flexibly to support systematic generalization. 
By incorporating such modules, diffusion-based DRL extends unsupervised disentanglement into a new class of generative modeling, where latent factors are not only identifiable and controllable but also embedded within a stable and highly expressive generative process—richly aligned with the goals of modular data representation without direct supervision.

Modular learning without modular supervision relies on clever objective functions and implicit cues to tease apart the underlying factors. Techniques enforcing statistical independence in the latent space have shown success on synthetic datasets, and they provide a foundation for disentangling factors of variation in an unsupervised manner. However, due to the ambiguities in purely unsupervised learning, practical implementations often incorporate mild forms of supervision or inductive biases (such as group data or known symmetries) to consistently attain interpretable, factorized representations. The trade-off is that without explicit supervision, the learned factors might not always align with human-labeled semantics, but they still offer a powerful form of modular representation that can be exploited for downstream tasks like reasoning, control, and generative manipulation.

\section{Modular Model Optimization}
\label{sec:modular_model}
While modular data representation focuses on disentangling representations at the data level, modular model learning emphasizes decomposing the architecture itself into specialized components. In this paradigm, a complex task is addressed not by a monolithic network but by a collection of modules $\{M_k\}_{k=1}^K$, each potentially responsible for a different sub-function, and coordinated under an overall architecture $\mathcal{A}$. The central question is how these modules are defined and trained: in some cases, their roles are predefined and guided with supervision, whereas in others, the system must discover modular structures and functions end-to-end. We first review methods with explicit modular supervision before turning to those without.

\subsection{Modular Model Optimization with Modular Supervision}
\label{sec:modular_model_with}
Modular models with \textit{modular supervision} assume that each module $M_k$ is associated with a predefined sub-task or functionality (e.g.\ perception vs.\ reasoning). The architecture $\mathcal{A}$ is composed of modules arranged to solve the overall task, and importantly, the role of each module is known \textit{a priori} and can be individually guided during training. This yields networks that are more interpretable and controllable, since each module’s output can be linked to a semantic function by design. We survey two major lines of work in this setting: neural \textbf{Modular Networks} with predefined structures, and \textbf{Neural-Symbolic Learning} where neural modules interface with symbolic reasoning under supervision.


\subsubsection{Modular Network}
\label{sec:modular_network}
Early Neural Module Networks (NMNs) exemplify the modular supervised approach. Andreas \textit{et al.}~\cite{Andreas2016Neural} first introduced Neural Module Networks (NMNs) as a compositional approach for visual question answering. Their model uses a semantic parser to decompose a complex question into sub-questions and dynamically assembles a neural network from a library of modules specialized for each sub-task. For example, given ``What color is the dog?'', the system might instantiate a sequence of modules {Find[dog]} $\to$ {Describe[color]}, where each module is designed to perform a specific function (e.g.\ finding an object or classifying its color). All modules are jointly trained, but each has an architecture aligned with its semantic role, such that intermediate outputs are interpretable (e.g.\ {Find[dog]} produces an attention heatmap over image regions likely containing a dog). This design exploits the linguistic substructure of questions to instantiate a tailored network on the fly. As a result, the reasoning process becomes more transparent: each module’s output can be examined rather than being an opaque vector in a monolithic network.

Subsequent work refined this paradigm. Johnson \textit{et al.}~\cite{Johnson2017Inferring} proposed to generate an explicit modular layout (a ``program'') from the question, instead of relying on a fixed parser. Their model consists of a {program generator} that parses the natural language into a sequence of predefined operations, and an {execution engine} that implements these operations with neural modules. For instance, the question is translated into a program using a fixed set of module types ({Find}, {Filter}, {Count}, {Compare}, etc.), which is then executed step-by-step by corresponding network modules. Both components are end-to-end trainable: the program generator is trained with supervision signals (and refined with REINFORCE to handle discrete decisions), while the execution engine modules learn their parameters via backpropagation. Importantly, the generated program serves as an interpretable reasoning trace for each question, detailing the sequence of reasoning steps the model took to arrive at its answer.

Mascharka \textit{et al.}~\cite{Mascharka2018Quantitative} further demonstrated that one can attain both high performance and interpretability by carefully designing modular networks. They developed the {Transparency by Design network (TbD-net)}, which executes CLEVR questions using a fixed layout of attention-based modules corresponding to the question’s functional program. By constraining each module to produce human-interpretable outputs (e.g.\ explicit attention masks passed between modules), TbD-net closed the gap between transparent models and prior black-box models. Each intermediate result (attention heatmaps, attribute labels, etc.) aligns with a reasoning step that a person could follow, enabling unparalleled diagnostic insight into the model’s behavior. This showed that modular VQA models can be built ``by design'' to be interpretable at each step without sacrificing accuracy.

Other researchers tackled the challenge of learning module layouts automatically to reduce dependence on hand-crafted parsers or program annotations. {End-to-End NMNs (N2NMN)}~\cite{Hu2017EndToEnd} learn to predict network layouts for each question without the aid of a fixed parser or layout supervision beyond the final answer. Their approach uses a sequence-to-sequence {layout policy} (a recurrent network) that dynamically decides which module to apply next, guided by the question. This policy is trained via a mix of imitation learning (using a small set of expert demonstration programs) and reinforcement learning, along with joint training of the modules’ parameters using the VQA loss. The result was a model that discovers question-specific network architectures on its own. 
In the follow-up work, they~\cite{Hu2018Stack} introduced the {Stack Neural Module Network (Stack-NMN)}, which further eliminates the need for any layout annotations or reinforcement-based training by making the composition process fully differentiable. The Stack-NMN employs a differentiable stack data structure to compose modules in sequence: instead of outputting a hard layout, the model’s controller continuously pushes and pops module outputs on a neural stack, effectively encoding the program execution in the stack’s state. This ``soft'' layout approach allows the entire reasoning process to be trained end-to-end with simple backpropagation, without any discrete layout decisions. Notably, Stack-NMN achieved accuracy on par with the best modular approaches on CLEVR, {without} using any ground-truth programs or rule-based parsers. Its outputs remained interpretable, and the authors demonstrated the method’s flexibility by applying the same set of modules to related tasks via the shared stack mechanism. In summary, Stack-NMN showed that learned layouts can match the performance of hand-designed ones, and it introduced a novel continuous relaxation for neural program execution that improved both accuracy and interpretability.

Collectively, these modular networks illustrate the power of injecting compositional structure into deep learning. By breaking down complex tasks into sub-tasks handled by dedicated components, NMNs achieved strong results on challenging reasoning benchmarks and provided human-readable reasoning traces. However, a key trade-off emerges: this interpretability often relies on prior knowledge or annotations to define the module structure. Early NMNs needed a syntactic question parser or functional program annotations; even the end-to-end variants assumed a fixed set of module types and some guidance to learn layouts. As Pahuja \textit{et al.}~\cite{Pahuja2019Structure} point out, ``a fundamental limitation of these approaches is that the modules need to be hand-specified, which becomes problematic when one has limited knowledge of the kinds of reasoning required for an open-domain task. In more unconstrained settings (e.g. \ real-image VQA with free-form questions), we often lack a well-defined grammar of operations, and questions may require novel combinations of reasoning skills. It is therefore challenging to extend NMNs directly to such settings—the model might not know what modules it needs, or how to parse an arbitrary question into a meaningful layout, without additional supervision.

Recent research has started addressing these challenges. One direction is to make the modules themselves more adaptive or learnable. MSR-ViR~\cite{songmodularized}  integrates a modular network into Multimodal LLMs for VideoQA. 
Their framework decomposes complex questions into sub-questions via a Question Parser and follows tree-structured policies to invoke temporal and spatial grounding modules. 
This modular reasoning process provides explicit reasoning paths and visual evidence for answers, significantly improving interpretability and transparency. Pahuja \textit{et al.}~\cite{Pahuja2019Structure} proposed a {Learnable Neural Module Network} that not only learns the layout but also learns the internal structure of each module in a data-driven way. Their approach uses differentiable architecture search to discover the computation performed inside each module, given only the final task loss, thereby removing the need for hand-engineering module architectures. Another line of work has integrated {neuro-symbolic} methods: for instance, Yi \textit{et al.}~\cite{Yi2018NeuralSymbolic} introduced a system that first converts the image into a structured scene graph and the question into a logical program, then executes the program on the scene graph with symbolic operations. On the other end of the spectrum, contemporary end-to-end models like FiLM and MAC networks have shown that one can obtain strong VQA performance without explicit modules by learning implicit reasoning circuits. These models use recurrent attention or feature-wise modulations to achieve compositional reasoning in a homogeneous architecture, sacrificing some interpretability for the sake of simplicity and ease of training. The evolution of NMNs thus highlights a continuum between structured, interpretable reasoning and unstructured, purely learned reasoning. The trend in recent years is to seek a middle ground: maintaining the transparent, modular nature of the reasoning process, while reducing the need for hand-crafted structure or supervision.

\subsubsection{Modular Models with Symbolic Constraints}
\label{sec:modular_model_symbolic_constraints}

While modular networks emphasize architectural decomposition by assigning different neural modules to specialized sub-functions, another complementary approach introduces modularity through symbolic constraints. Instead of relying solely on structural design, these methods embed symbolic knowledge directly into neural learning, thereby constraining modules to operate in a logically consistent manner. From the perspective of Modular Machine Learning (MML), this represents a second form of supervision: symbolic priors shape how representations are learned, how modules interact, and how reasoning unfolds. Such integration bridges the flexibility of neural approximation with the rigor of symbolic reasoning, aligning the data-driven and rule-based paradigms within a unified modular framework.

\begin{table}[ht]
\centering
\caption{Representative neural-symbolic learning methods with Symbolic Constraints on Neural Models. 
This paradigm integrates symbolic priors or logical rules directly into neural architectures, enforcing structured reasoning during learning. 
Abbreviations: Img. Cls. = Image Classification; Prog. Syn. \& Code = Program Synthesis \& Code Understanding; Img. Anal. = Image Analysis; 
Causal Learn. = Causal Learning; VQA = Visual Question Answering; Vid. Gen. = Video Generation.}
\vspace{-2mm}

\begin{adjustbox}{max width=0.4\textwidth}
\begin{tabular}{|l|l|l|}
\hline
\textbf{Method} & \textbf{Reasoning Capability} & \textbf{Applications} \\
\hline
LTN~\cite{SerafiniGarcez2016LTN} & Deductive  & Img. Cls. \\
Semantic Loss~\cite{XuEtAl2018SemanticLoss} & Deductive  & Img. Cls. \\
DL2~\cite{FischerEtAl2019DL2} & Deductive  & Img. Cls. \\
DILP~\cite{ShakarianEtAl2023DILP} & Inductive  & Prog. Syn. \& Code \\
MIL~\cite{MuggletonEtAl2014MIL} & Inductive  & Prog. Syn. \& Code \\
Forth~\cite{BosnjakEtAl2016DifferentiableForth} & Inductive  & Prog. Syn. \& Code \\
DeepProbLog~\cite{Manhaeve2021DeepProbLog} & Deductive  & Prog. Syn. \& Code \\
NeurASP~\cite{YangEtAl2020NeurASP} & Deductive  & Img. Anal. \\
NSCL~\cite{MaoEtAl2019NSCL} & Deductive  & VQA \\
PVR~\cite{li2019perceptual} & Inductive  & VQA \\
DSTN~\cite{qian2022dynamic} & Inductive  & VQA \\
Modular-Cam~\cite{pan2025modular} & Inductive  & Vid. Gen. \\
CausalVAE~\cite{YangEtAl2020CausalVAE} & Causal / Counterfactual  & Causal Learn. \\
CGN~\cite{SauerGeiger2021CGN} & Causal / Counterfactual  & Causal Learn. \\
\hline
\end{tabular}
\end{adjustbox}
\vspace{-2mm}
\end{table}

One direction incorporates symbolic rules directly into the training objective of neural models. Logic Tensor Networks (LTN)~\cite{SerafiniGarcez2016LTN} embed first-order logic formulas as differentiable terms in the loss function, ensuring that neural predictions simultaneously minimize empirical risk and logical inconsistency. Semantic Loss~\cite{XuEtAl2018SemanticLoss} takes a similar approach, encoding domain-specific symbolic knowledge into a loss term that penalizes violations of constraints. DL2~\cite{FischerEtAl2019DL2} generalizes this principle by defining a differentiable framework for arbitrary logical conditions, enabling flexible integration of symbolic rules into gradient-based optimization. From the MML perspective, these methods treat symbolic constraints as {modular priors}: they modularize the latent representation space so that features align with symbolic rules, embedding logical structure into neural slots. A limitation is scalability, since encoding large or complex rule sets often leads to computational intractability and difficulties in generalization beyond well-defined logical domains.

Beyond regularization, symbolic supervision can guide explicit decomposition of tasks into modular sub-components. Perceptual Visual Reasoning (PVR)~\cite{li2019perceptual} provides an early example, decomposing visual reasoning tasks into perceptual and logical modules and propagating supervision across them. Symbolic-like operators such as AND and OR are instantiated as neural components, producing outputs that reflect compositional reasoning while preserving end-to-end trainability. The Dynamic Spatio-Temporal Modular Network (DSTN)~\cite{qian2022dynamic} extends this principle to video question answering: it applies supervision to spatial, temporal, and logical modules, dynamically assembling tree-structured reasoning paths over objects, actions, relations, and temporal orders. Each specialized module outputs interpretable intermediate representations, illustrating how symbolic decomposition can yield human-understandable reasoning traces. A further extension is Modular-Cam~\cite{pan2025modular}, which leverages large language models (LLMs) to parse natural language instructions into structured scene graphs and camera operations. These are then executed by specialized neural operators (e.g., \textit{ZoomIn}, \textit{PanLeft}), each acting as a symbolic-to-neural translation module. This design demonstrates how symbolic-level decomposition can be realized in multimodal generation, supporting controllability and interpretability in complex video synthesis.

Symbolic supervision has also been used in reasoning-focused applications, where the primary goal is to induce or enforce logical structures. Inductive Logic Programming (ILP) and its neural-symbolic extensions such as DILP~\cite{ShakarianEtAl2023DILP} and MIL~\cite{MuggletonEtAl2014MIL} embed ILP principles into differentiable frameworks, enabling neural networks to induce symbolic rules from training examples. Deductive reasoning has been addressed by systems such as DeepProbLog~\cite{Manhaeve2021DeepProbLog}, which integrates probabilistic logic programming with neural predicates; NeurASP~\cite{YangEtAl2020NeurASP}, which couples deep learning with answer set programming; and Differentiable Forth~\cite{BosnjakEtAl2016DifferentiableForth}, which models program execution as a differentiable process. These systems explicitly maintain symbolic supervision to guarantee logical validity, while neural components provide robustness to noisy perception and scalability in learning. Causal and counterfactual reasoning form another sub-area: CausalVAE~\cite{YangEtAl2020CausalVAE} integrates structural causal models into variational autoencoders, aligning latent modules with causal factors, while Counterfactual Generative Networks (CGN)~\cite{SauerGeiger2021CGN} generate counterfactual data by intervening on causal variables in the latent space. These methods supervise modules with causal knowledge, enforcing disentanglement and interpretability under explicit causal structures.

Overall, modular models with symbolic constraints demonstrate how symbolic knowledge can be operationalized as a supervisory signal to guide neural architectures. By embedding logical rules as differentiable constraints, decomposing complex tasks into symbolic sub-modules, or supervising reasoning and causal inference, these methods modularize neural models in ways that enhance interpretability, functional specialization, and compositional reasoning. While challenges such as scalability and integration with large-scale, noisy data remain, symbolic constraints provide a concrete mechanism to embed modular structure into neural learning, complementing architectural modular networks and advancing the broader goals of MML.

\subsection{Modular Model Optimization without Modular Supervision}
\label{sec:modular_model_without}
In the absence of modular supervision, a model must learn both the architecture $\mathcal{A}$ and the specialization of each module end-to-end from the overall task objective alone. There is no external annotation or constraint that, for example, “module 1 must detect objects” or “module 2 must perform reasoning”. Instead, the system is given a search space $\mathcal{S}$ of possible module configurations and must discover an effective design by optimizing $\mathcal{L}_{task}$. Two prominent directions fall in this category: Neural Architecture Search (NAS), which automatically designs neural network topologies (often effectively discovering modular structures) without hand-crafted module roles; and Neural-Symbolic Learning approaches, where the model might internally develop modular or symbolic behaviors without being told what those should be. The lack of explicit per-module supervision offers greater flexibility and potential for discovering novel architectures, but often at the cost of interpretability, since the learned modules are not guaranteed to correspond to human-understandable functions. 

\subsubsection{Neural Architecture Search}
\label{sec:nas}

Neural architecture search (NAS) aims to automatically design neural networks by exploring a predefined search space of candidate models to maximize task performance, which is particularly valuable when manual design is costly~\cite{elsken2019neural}. Without modular supervision, NAS performs end-to-end optimization to identify the optimal parameters for each sub-module or neural network. In general, NAS methods comprise three components: (1) the \textbf{search space}, which defines the set of candidate architectures; (2) the \textbf{search strategy}, which dictates how to explore the space; and (3) the \textbf{evaluation strategy}, which assesses the performance of each candidate~\cite{elsken2019neural}. We organize our survey of NAS literature around these components, whose representative methods are summarized in Table~\ref{tab:nas}.

\begin{table}[ht]
\centering
\caption{Summary of representative NAS methods. Abbreviations for Search Strategy (ST): 
RL = Reinforcement Learning, EA = Evolutionary Algorithm, BO = Bayesian Optimization, Con. = Continuous.}
\vspace{-2mm}
\label{tab:nas}
\begin{adjustbox}{max width=0.4\textwidth}
\begin{tabular}{|l|l|l|l|}
\hline
\textbf{Method} & \textbf{Search Space} & \textbf{ST} & \textbf{Evaluation Strategy} \\
\hline
NASNet~\cite{zoph2018learning} & Cell-based & RL & Proxy Training \\
ENAS~\cite{pham2018enas} & Cell-based & RL & Weight Sharing \\
MnasNet~\cite{tan2019mnasnet} & Efficient & RL & Proxy Training \\
AmoebaNet~\cite{real2019regularized} & Cell-based & EA & Proxy Training \\
Genetic CNN~\cite{xie2017genetic} & Cell-based & EA & Proxy Training \\
NASBOT~\cite{kandasamy2018nasbot} & Efficient & BO & Early Stopping \\
BANANAS~\cite{white2021bananas} & Cell-based & BO & Early Stopping \\
DARTS~\cite{liu2018darts} & Cell-based & Con. & Weight Sharing \\
SNAS~\cite{xie2019snas} & Cell-based & Con. & Weight Sharing \\
GDAS~\cite{dong2019gdas} & Cell-based & Con. & Weight Sharing \\
P-DARTS~\cite{chen2019progressive} & Cell-based & Con. & Proxy Training \\
DARTS-~\cite{chu2021darts} & Cell-based & Con. & Weight Sharing \\
PC-DARTS~\cite{xu2020pcdarts} & Cell-based & Con. & Weight Sharing \\
ProxylessNAS~\cite{cai2018proxylessnas} & Cell-based & Con. & Weight Sharing \\
EcoNAS~\cite{zhou2020econas} & Cell-based & Con. & Proxy Training \\
Hyperband~\cite{li2018hyperband} & Cell-based & Con. & Early Stopping \\
NASWOT~\cite{mellor2021naswot} & Cell-based & Con. & Zero-Cost Proxies \\
Zero-Cost NAS~\cite{abdelfattah2021zero} & Cell-based & Con. & Zero-Cost Proxies \\
\hline
\end{tabular}
\end{adjustbox}
\end{table}

The first component, the search space, defines which neural architectures are considered during NAS~\cite{elsken2019neural}. It typically has two aspects: an \emph{operation space} (the set of allowable operations or layer types, such as convolutions, pooling, etc., along with their hyperparameters) and a \emph{connection space} (how those operations can be arranged or connected). A well-designed search space can introduce helpful inductive biases that simplify the search and improve final performance. NAS search spaces are often characterized at two levels: \emph{macro}-architecture (the entire network topology) vs. \emph{micro}-architecture (reusable modules or cells). Early NAS works focus on the macro level: for example, Zoph \textit{et al.} uses a controller RNN to sequentially generate a whole network layer by layer~\cite{zoph2016neural}. Their search space is a sequence of layers with various convolutional options, including possible skip connections. Although effective, such a large macro search space can be extremely expensive to explore. To address this, differentiable methods such as DARTS~\cite{liu2018darts} introduced a continuous relaxation of architecture parameters, enabling gradient-based optimization in large spaces. A notable development in this direction is the cell-based search space, which breaks the architecture into smaller, repeatable modules called \emph{cells}. Instead of searching for an entire network, NAS searches only for the structure of these cells (represented as directed acyclic graphs), which are then stacked or repeated to form the final network~\cite{zoph2018learning}. This modular design dramatically reduces the search space while maintaining the capacity to build very deep models. Zoph \textit{et al.}~\cite{zoph2018learning} introduced NASNet, which learns two types of cells (normal and reduction) on CIFAR-10 and transfers them to larger datasets like ImageNet. Other works, such as AmoebaNet~\cite{real2019regularized} and DARTS~\cite{liu2018darts}, also employ cell-based spaces. ENAS~\cite{pham2018enas} further accelerated the search by parameter sharing within a supernet, reducing computation time by orders of magnitude. Beyond modularity, many works also focus on efficiency: since a naive search space can be prohibitively large, constraints and progressive strategies are often introduced. For instance, Curriculum NAS (CNAS)~\cite{guo2020breaking} starts with a small subspace and gradually expands it as knowledge accumulates, while Progressive NAS (PNAS)~\cite{liu2018progressive} incrementally adds complexity using surrogate models to prune poor candidates. Another efficiency strategy is weight sharing or one-shot NAS, where a single over-parameterized supernet encompasses all candidate architectures, and each architecture is a subgraph. ENAS~\cite{pham2018enas} and ProxylessNAS~\cite{cai2019proxylessnas} are prime examples. Furthermore, some search spaces are hardware-aware, including only lightweight operations (e.g., depthwise separable convolutions) to meet latency and memory constraints~\cite{wu2019fbnet}.

Having established the role of search spaces, we now turn to the second component: search strategy. NAS employs various strategies to explore the vast architecture space, which can be broadly categorized into discrete search methods and continuous (differentiable) search methods. Discrete search methods involve explicitly sampling and evaluating architectures in an iterative loop. They do not assume a differentiable search space; instead, they rely on techniques such as reinforcement learning, evolutionary algorithms, or Bayesian optimization to guide the search. These approaches are conceptually straightforward and highly general (they can optimize any architecture representation), but they often entail high computational costs because each candidate architecture typically needs to be trained (at least partially) from scratch. Reinforcement learning (RL)-based NAS uses a controller (e.g., an RNN) to generate architectures sequentially. Zoph and Le~\cite{zoph2016neural} pioneered this approach, training a controller to sample convolutional networks. Subsequent works improved efficiency: NASNet~\cite{zoph2018learning} uses transferable cells; ENAS~\cite{pham2018enas} shares parameters in a supernet; MnasNet~\cite{tan2019mnasnet} adds latency constraints. RL methods are flexible but computationally heavy. Evolutionary algorithms (EA) offer an alternative, treating NAS as a population-based optimization. Real \textit{et al.}~\cite{real2017large,real2019regularized} evolved architectures using mutation and selection, introducing aging evolution (AmoebaNet). Genetic CNN~\cite{xie2017genetic} and NEAT~\cite{stanley2002neat} are other examples. EA is parallelizable and robust, often combined with weight inheritance to reduce cost. Another family of discrete approaches is Bayesian optimization (BO), which builds a surrogate model of architecture performance to guide search. NASBOT~\cite{kandasamy2018nasbot} introduces a kernel on architectures; BANANAS~\cite{white2021bananas} uses path encodings and ensembles. BO is sample-efficient but needs careful encoding of architectures.

In contrast, continuous (differentiable) search methods redefine NAS as a learning problem solvable by gradient-based optimization, dramatically speeding up the search. DARTS~\cite{liu2018darts} relaxes discrete operation choices into continuous variables $\alpha$ (weights for candidate operations). A supernet containing all candidate operations is trained in a bi-level fashion: network weights are optimized on training data while architecture parameters $\alpha$ are learned on validation data. By alternating these updates, DARTS learns a cell architecture, which is then discretized by selecting the highest-weighted operation on each edge. Despite its efficiency, vanilla DARTS can suffer from performance collapse, converging to degenerate architectures dominated by trivial operations (e.g., skip connections). This occurs because skips propagate gradients without attenuation, making them overly attractive during optimization. As a result, unconstrained gradient descent often yields unstable searches and poor architectures when retrained from scratch. Several DARTS variants have been proposed to stabilize the search and improve results. SNAS~\cite{xie2019snas} adds stochasticity by treating $\alpha$ as parameters of a categorical distribution and sampling one operation per edge each forward pass. This aligns gradient optimization with eventual discrete choices and prevents any operation from dominating too early. P-DARTS~\cite{chen2019progressive} progressively increases network depth during search to bridge the “depth gap” between search and evaluation. DARTS typically searches on a shallow proxy network; P-DARTS gradually deepens the supernet and prunes weaker operations in stages. This ensures the discovered cell works well in deeper networks and yields competitive ImageNet results with a fast search schedule. 
ProxylessNAS~\cite{cai2018proxylessnas} searches directly on the target task and incorporates device-specific latency into the objective. It gates one path at a time and optimizes for both accuracy and inference speed. The result is models tailored to real-device performance, showing that differentiable NAS can handle multi-objective optimization.
In summary, continuous differentiable NAS methods like DARTS and its variants revolutionize NAS by dramatically reducing search costs and enabling effective searches on modest compute budgets. While one-shot weight-sharing approaches introduce some biases (e.g., certain operations dominating early), various improvements mitigate these issues. Consequently, differentiable NAS remains a dominant approach that consistently yields state-of-the-art models, and ongoing research continues to close the gap between the search phase and final model performance.

The final component of NAS is the evaluation strategy, which addresses the challenge of estimating an architecture’s performance without fully training it to convergence. Since full training is often prohibitively slow and resource-intensive, several acceleration techniques have been developed.
 Below, we survey four key approaches for speeding up NAS evaluation in an accessible manner, introducing the core ideas and benefits of each.
One widely adopted approach is proxy training. Instead of fully training each candidate on the target task, models are evaluated on smaller datasets, for fewer epochs, or with reduced-size networks to approximate performance~\cite{zoph2016neural,zoph2018learning}. This greatly reduces cost by providing a quick but coarse signal about which architectures are promising. However, the fidelity of proxies is not guaranteed; EcoNAS~\cite{zhou2020econas} systematically studied this and found only moderate correlation between proxy and final performance. Still, proxy tasks are widely used as a first filter, as they allow practitioners to test many candidates at a fraction of the full cost.
Another important technique is early stopping, which aims to terminate poorly performing runs before completion, thereby saving time and resources. The intuition is that models showing weak performance in early epochs are unlikely to become competitive later. Approaches in this category include learning-curve extrapolation~\cite{domhan2015lc}, which predicts final accuracy from partial training curves, and adaptive resource allocation such as Hyperband~\cite{li2018hyperband}, which trains many candidates briefly and progressively allocates more epochs to the top performers. These strategies can dramatically reduce total compute by focusing effort only on likely winners.
A third family of methods is based on weight sharing, also known as one-shot NAS. Instead of training each architecture independently, a large “supernet’’ is constructed that contains all candidate operations~\cite{pham2018enas,brock2018smash,bender2018oneshot}. During training, sub-architectures inherit weights from shared parameters, meaning that evaluation requires only a forward pass. This yields orders-of-magnitude speedups and enables exploration of vast search spaces. While this approach can introduce bias—since shared weights may not reflect standalone performance—it has become a key technique in efficient NAS due to its scalability.
Finally, zero-cost proxies have recently been proposed to estimate architecture quality without training. These methods compute lightweight metrics at initialization, such as gradient sensitivity, synaptic flow, or activation diversity (NASWOT~\cite{mellor2021naswot}). Abdelfattah \textit{et al.}~\cite{abdelfattah2021zero} demonstrated that such scores can correlate surprisingly well with final accuracy, offering a near-instant signal to rank thousands of architectures. Although noisier than trained estimates, zero-cost metrics are extremely cheap and are increasingly combined with other strategies for fast architecture screening.

\subsubsection{Inductive Modularization}
\label{sec:modular_without_symbolic_supervision}

Moving beyond Neural Architecture Search (NAS) for discrete module structures, an alternative paradigm is to induce modularity implicitly through end-to-end differentiable optimization. In these approaches, the model is not explicitly supervised with symbolic module labels or structures; instead, modular behavior emerges by embedding symbolic elements into neural representations and by learning neural analogues of reasoning operations. This allows a network to learn to compose sub-components (modules) through gradient-based training, rather than by enumerating architectures. For example, rather than searching over discrete architectures as in NAS, one can embed symbolic relations or operations directly into continuous vectors or neural operators, effectively creating ``soft'' modules that the model can reuse and compose. Such differentiable modular induction leverages techniques like knowledge graph embeddings, neural reasoning layers, and programmatic networks to achieve modularity without requiring hand-crafted symbolic supervision.

\begin{table}[ht]
\centering
\caption{Representative neural-symbolic learning methods with Neural Models for Symbolic Representation. 
This paradigm encodes symbolic entities, relations, or logic into neural embeddings, allowing symbolic structures to be processed by deep models. 
Abbreviations: KG = Knowledge Graph; KB = Knowledge Base; Vis. Prog. Ind. = Visual Program Induction.}
\begin{adjustbox}{max width=0.9\textwidth}
\begin{tabular}{|l|l|l|}
\hline
\textbf{Method} & \textbf{Reasoning Capability} & \textbf{Applications} \\
\hline
TransE~\cite{BordesUGWY13TransE} & Inductive  & KG \\
RotatE~\cite{SunDengNieTang2019RotatE} & Inductive  & KG \\
NTP~\cite{RocktaschelRiedel2017NTP} & Deductive  & KB \\
FM~\cite{duan2022parametric} & Inductive  & Vis. Prog. Ind. \\
LNN~\cite{RiegelEtAl2020LNN} & Deductive  & Symb. Reas.  \\
\hline
\end{tabular}
\end{adjustbox}
\end{table}

A prominent class of approaches embeds symbolic entities and relations as vectors or transformations in a latent space, enabling relational reasoning via algebraic operations in that space. In knowledge graph embedding (KGE) methods such as TransE~\cite{conf/nips/BordesUGWY13} and RotatE~\cite{journals/corr/abs-1902-10197}, each entity and relation is represented by a vector (or complex vector) in the same continuous space. TransE enforces a compositional rule $\mathbf{h} + \mathbf{r} \approx \mathbf{t}$ for a triple $(h, r, t)$, while RotatE models each relation as a rotation in the complex plane. By casting symbolic relations into continuous operations, these embedding-based models can efficiently learn from large-scale knowledge graphs and generalize to predict missing links using simple distance or score functions. Within MML, these embeddings can be regarded as latent ``modules''---entities correspond to representation slots, relations correspond to parameterized transformations, and reasoning corresponds to chaining such modules. This yields scalability and flexibility, but the reasoning process is opaque and loses explicit logical interpretability.

To introduce more explicit reasoning operators, Neural Theorem Provers (NTPs)~\cite{RocktaschelRiedel2017NTP} combine embeddings with logic-inspired computation. An NTP builds a differentiable analogue of a logical prover: symbols are represented as embeddings, and proof steps (unifications, rule applications) are implemented by neural modules operating on these embeddings. The proving process becomes a recursively constructed neural network whose success score is differentiable with respect to symbol embeddings. This enables gradient descent to jointly train embeddings and rule parameters. Through end-to-end training, NTPs implicitly learn modular components: they place analogous symbols close in the vector space and can induce simple logic rules from data. Thus, the reasoning operators of an NTP---differentiable unification and logic inference---function as soft modules for reasoning. Compared to supervised symbolic methods, these modules are fuzzy and approximate, but they balance neural generalization with logic-like compositionality.

Another direction achieves modularity by learning function-like or logic-like modules within neural networks, especially for tasks requiring compositional reasoning. Function Modularization(FM)~\cite{duan2022parametric} models each parametric function and its parameters as a self-contained neural module, trained without symbolic labels. A hierarchical heterogeneous Monte Carlo Tree Search (H2MCTS) explores program compositions during training, providing supervision signals for the neural functions. Through this process, the model learns a reusable library of neural subroutines that can be dynamically assembled to solve novel visual queries. Logical Neural Networks (LNNs)~\cite{RiegelEtAl2020LNN} adopt a similar philosophy by embedding logical connectives (e.g., AND, OR) as differentiable operators. Each neuron corresponds to a component of a logical formula, and rule confidences are learned via gradient descent. This design yields interpretable modular components aligned with logical structure, while remaining trainable end-to-end. In both PVPI and LNNs, modularity is discovered implicitly: the system learns to approximate functional or logical operators as neural modules guided only by task-level objectives.

A central challenge is coordinating module learning with module selection and composition. Some systems employ \emph{soft selection}, using gating or attention mechanisms to blend candidate modules, which allows gradients to flow and modules to specialize. Mixture-of-experts models are a typical example~\cite{masoudnia2014mixture}, where a gating network selects experts (modules) adaptively. Other systems rely on explicit \emph{search over compositions}, as in FM~\cite{duan2022parametric} where Monte Carlo tree search discovers effective module sequences. Hybrid approaches combine both relaxation and search. These training strategies yield specialized neural modules and controllers that know how to compose them. Applications are broad: in knowledge graph reasoning, embeddings and NTPs enable scalable multi-hop inference; in visual question answering, program-induction methods compose perceptual and reasoning modules to answer complex queries; and in multimodal generation, modular inductive models coordinate consistency across modalities via structured sub-components.

Overall, unsupervised modular induction shows how modularity can emerge from data without explicit symbolic supervision. Embedding-based methods modularize reasoning into vectors and transformations, while program-inductive methods construct functional operators as implicit neural modules. These approaches scale to large, noisy datasets and adapt flexibly across tasks, but they typically lack strong interpretability and guarantees of logical soundness. In comparison, symbolically supervised methods produce modules that are more interpretable and better aligned with human-understandable symbolic structures, but require annotated supervision and do not scale as easily. Together, these paradigms illustrate complementary strategies in MML: explicit supervision enforces clear modular semantics, whereas unsupervised induction lets modularity emerge autonomously, trading transparency for scalability.

\section{The Key Significance of MML for LLMs}
\label{sec:mml-significance}

~~~\textbf{Enable Counterfactual Reasoning.}  
One of the most profound limitations of current LLMs is their inability to effectively reason about counterfactuals~\cite{hoch1985counterfactual}. Counterfactual reasoning explores hypothetical ``what if'' scenarios, such as estimating outcomes had a patient received a different treatment. While LLMs excel at pattern-based prediction, they remain ill-suited for tasks demanding logical or causal consistency. MML addresses this gap by integrating modular reasoning systems grounded in rules or causal graphs, thereby enabling coherent counterfactual hypotheses~\cite{dong2019neural}. By decoupling perception from reasoning, MML mitigates statistical biases of neural networks and enforces causal constraints, offering a robust foundation for counterfactual reasoning within LLMs~\cite{cambria2024xai}.

\textbf{Conduct Rule-Based Reasoning.} One of the main challenges of LLMs is their reliance on probabilistic text generation, which can lead to factually incorrect outputs~\cite{zeng2023logical}. MML enhances LLMs by introducing structured rule-based reasoning mechanisms that ensure factual consistency. Modules within MML frameworks can extract rules from LLM-generated content and validate them against external knowledge bases, ontologies, or predefined constraints.
For example, in the medical domain, an LLM could synthesize clinical guidelines derived from scientific literature, while an MML-integrated modular reasoning module can cross-check these recommendations against structured knowledge, identifying inconsistencies or inaccuracies. This approach enhances the reliability of LLM outputs, making them suitable for high-stakes applications where precision is critical. Additionally, by continuously updating its rule base, MML ensures that LLMs remain aligned with up-to-date domain knowledge, thereby reducing the risk of outdated or biased recommendations.

\textbf{Unify Perception and Reasoning.} The integration of MML and LLM into a unified framework bridges the gap between perception and reasoning, offering a scalable and adaptable AI paradigm. By leveraging MML’s modular nature, this unified system can efficiently handle complex multi-modal data, combining LLM-driven perception with modular reasoning to enable comprehensive decision-making.
One of the key benefits of this approach lies in its flexibility. While LLMs can be fine-tuned for domain-specific applications, MML’s modular design ensures that different reasoning modules can be updated independently. For example, in an autonomous system, the perceptual module (LLM) can process sensor data and natural language commands, while the reasoning module (MML) ensures compliance with predefined safety constraints. This separation enhances system robustness and reduces the need for extensive retraining when adapting to new tasks.

\textbf{Eliminate Hallucinations.}  
Hallucinations, factually incorrect or inconsistent outputs, arise from LLMs’ probabilistic training and pose severe risks in domains such as healthcare and law. MML mitigates this by embedding modular reasoning components that enforce factuality through knowledge bases, ontologies, or rules. For instance, a medical module can cross-check LLM recommendations against clinical guidelines. MML also supports iterative refinement, where outputs (e.g., scientific summaries) are validated via domain-specific rules, ensuring logical consistency. Beyond post-processing, modular filters can dynamically monitor and correct outputs during generation, such as enforcing statutory compliance in legal drafting. These mechanisms collectively enhance reliability and make LLMs more trustworthy for high-stakes applications.

\textbf{Encourage Fairness and De-Biasing.}
The issue of fairness and bias in LLMs arises from the biases hidden in their training data. These biases can manifest as discriminatory behaviors in applications ranging from recruitment to lending~\cite{li2023survey}. While traditional methods focus on mitigating bias at the data preprocessing or model training stage, MML introduces a novel paradigm by embedding fairness constraints directly into the reasoning process.
MML-based modules within the framework can explicitly define fairness criteria, ensuring that decisions align with ethical guidelines and societal norms. For instance, in a job recommendation system, modular rules could enforce demographic parity by evaluating whether candidates from different groups receive equitable recommendations. Similarly, in lending scenarios, modular reasoning could validate creditworthiness decisions to prevent discrimination based on race or gender.
In addition to enforcing fairness, MML can further facilitate bias detection and correction. Neural modules can preprocess large datasets to identify patterns indicative of bias, while modular reasoning can analyze these patterns to infer the underlying causations. This dual capability enables a comprehensive approach to addressing bias, combining the scalability of neural networks with the interpretability of modular systems.
Moreover, the modularity of MML allows for the continuous evolution of fairness criteria. As societal norms change, human-understandable rules can be updated without retraining the entire system. This adaptability ensures that MML-enhanced LLMs remain aligned with contemporary ethical standards, setting a new benchmark for responsible AI development.

\textbf{Enforce Robustness.} 
Safety is a fundamental requirement for deploying AI systems in real-world scenarios, particularly in critical domains such as healthcare, finance, and autonomous systems. Traditional LLMs, while powerful, lack intrinsic mechanisms to ensure safety, making them susceptible to adversarial attacks, out-of-distribution scenarios, and catastrophic errors~\cite{kumar2023certifying, wei2023jailbroken}.
MML addresses these vulnerabilities by embedding safety constraints within MML-based modules. These modules act as ``guardrails", enforcing rules that prevent unsafe behaviors. For instance, in autonomous driving applications, modular systems could validate decisions made by LLMs against predefined safety protocols, such as maintaining a safe distance from other vehicles or adhering to traffic laws.
The robustness of MML-enhanced LLMs extends beyond safety constraints. MML-based modules can perform real-time anomaly detection by analyzing outputs to check inconsistencies or deviations from expected patterns. In the domain of cybersecurity, for example, modular reasoning could identify unusual network activity indicative of a potential breach, enabling proactive intervention.

\textbf{Enhance Interpretability.} 
The interpretability of AI systems is a pressing concern, particularly as they are increasingly deployed in domains where accountability is paramount~\cite{huang2023can}. Traditional LLMs, often described as ``black box", struggle to provide explanations for their outputs, limiting their adoption in critical applications.
MML offers a solution by integrating modular reasoning modules that generate interpretable outputs. These modules translate complex neural representations into human-readable logical explanations, bridging the gap between machine learning and human understanding. For instance, in a legal AI system, MML-based modules could explain how specific statutes and precedents influenced a recommendation, providing a clear audit trail for decisions.
The modular architecture of MML also facilitates domain-specific customization of explanations. By tailoring modular rules to the requirements of a particular domain, MML can generate contextually relevant explanations. In healthcare, for example, modular systems could elucidate the clinical pathways leading to a diagnosis, enabling practitioners to validate and refine their decisions.
Moreover, MML-enhanced LLMs support interactive explainability, allowing users to query and explore the reasoning behind outputs. For instance, in a financial application, users could ask why a loan application was rejected and receive a detailed explanation of the factors considered. This interactive capability not only enhances transparency but also fosters user trust and engagement.
Last but not least, interpretability can also facilitate trust in AI systems by providing a clear rationale for decisions. For instance, in medical diagnostics, modular systems could generate detailed explanations for treatment recommendations, enabling clinicians to verify their validity. This transparency is essential for trust through fostering confidence in AI systems, particularly in high-stakes applications.

Therefore, LLMs have demonstrated remarkable capabilities in processing unstructured data, understanding natural language, and capturing semantic nuances across various modalities. However, their limitations in reasoning, factual consistency, fairness, and interpretability necessitate an enhanced framework that integrates MML principles. MML provides a structured backbone that augments LLMs, ensuring logical coherence, adaptability, and robustness in real-world applications.
One of the critical advantages of MML is its modular architecture, which enables the decoupling of perception and reasoning. LLMs, with their superior ability to parse complex, high-dimensional data, can serve as the perceptual layer, converting raw inputs into structured representations. These structured representations can then be processed by dedicated reasoning modules within the MML framework, ensuring consistency, interpretability, and logical soundness. 
Of course, developing a unified framework requires addressing the technical challenge of integrating continuous representations produced by LLMs with discrete modular reasoning. Differentiable reasoning architectures, such as neuro-modular integration pipelines, facilitate seamless interaction between these components. These architectures approximate modular logic with continuous operations, enabling gradient-based optimization while preserving logical consistency.

The interaction between MML and LLMs leads to emergent properties that neither system can achieve independently. Inspired by cognitive dual-process theories~\cite{beevers2005cognitive}, this hybrid system enables intuitive, fast decision-making through LLMs while ensuring deliberate, logical reasoning via MML-based modules. This dual-layered approach significantly improves decision accuracy and adaptability across diverse applications.
In addition, the interplay between LLM-driven content generation and MML-based validation is able to foster a robust iterative process for hypothesis generation and verification. 
By structuring LLMs within the MML framework, AI systems can achieve enhanced robustness, interpretability, and adaptability, setting the foundation for the next generation of unified intelligent systems~\cite{wu2024next}.

\begin{figure*}[!htbp]
\centering
    \includegraphics[width=0.8\textwidth]{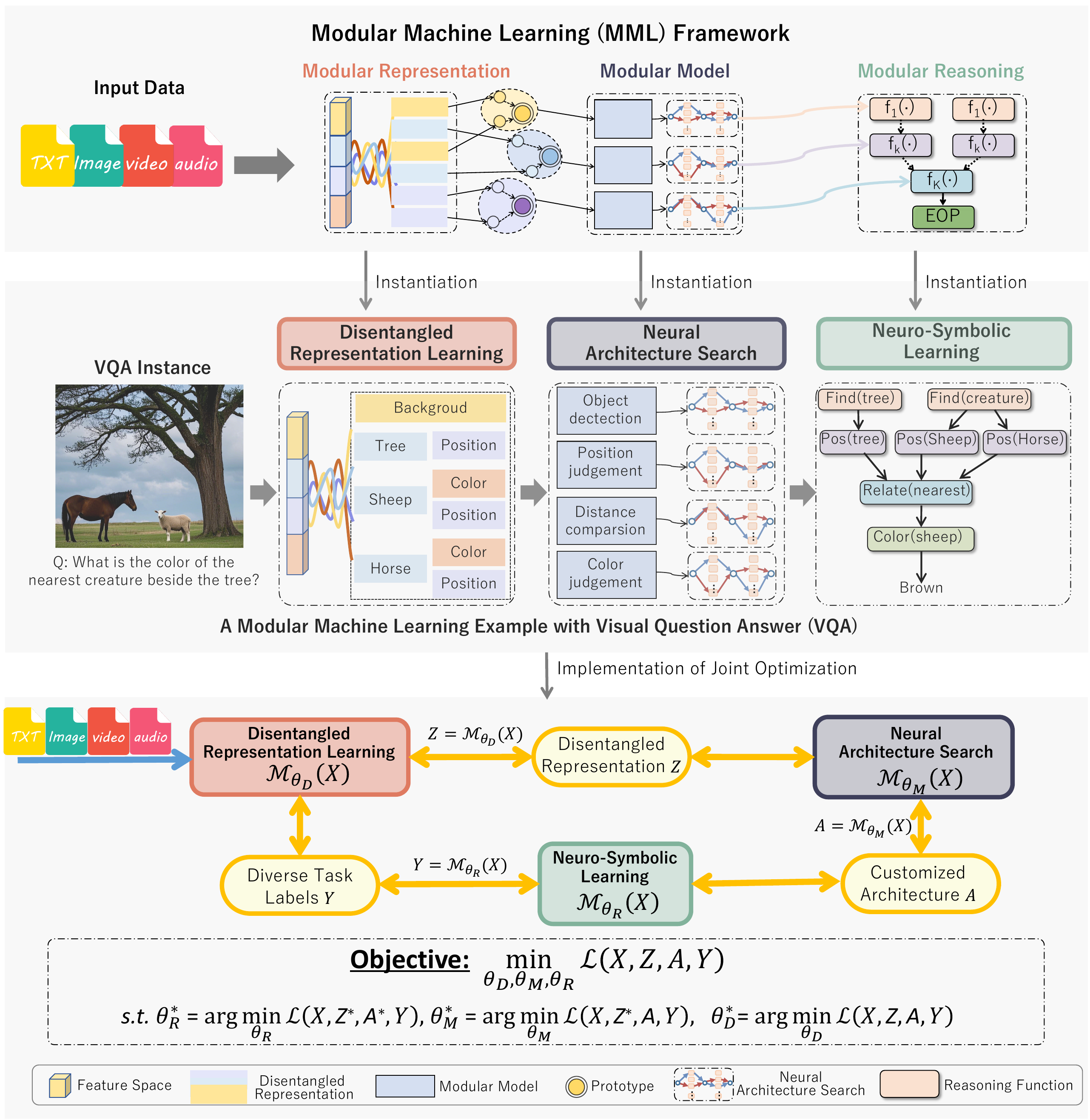}
\caption{
The proposed unified MML framework (Modular Representation, Modular Model, and Modular Reasoning) with a feasible implementation (Disentangled Representation Learning, Neural Architecture Search, and Neuro-Symbolic Learning) to solve a practical task. To maintain consistency with Fig.~\ref{fig:intro}, we again take a visual question answering (VQA) task as an example to illustrate the instantiation of MML. We utilize the VQA example to demonstrate how MML disentangles the input into modular components, processing through a sequential combination of disentangled representation learning, neural architecture search, and neuro-symbolic learning. Specifically, the disentangled representation learning component is responsible for extracting a set of disentangled features from the input data, which are subsequently fed into the neural architecture search component. The neural architecture search component then identifies the optimal architecture that best captures the disentangled representations. Following this, the neuro-symbolic learning component further refines the output through structured, logic-driven reasoning.
In the VQA example, the input image that contains a tree, a horse, and a sheep is processed by MML to extract disentangled representations indicating background, position, and color etc. Based on these disentangled representations, MML then discovers the optimal neural architecture to perform various tasks, including object detection, position judgment, distance comparison, and color identification. Finally, neuro-symbolic learning integrates the outputs in a logical reasoning manner to answer the question: ``What is the color of the nearest creature beside the tree?'' The answer, ``Brown'' correctly identifies the nearest creature (the sheep) and its color, using symbolic logic to establish the relationship between the entities.
Additionally, we formally present the joint optimization process at the bottom of the figure, which involves the simultaneous optimization of the three modules: disentangled representation learning, neural architecture search, and neuro-symbolic learning. The objective function aims to minimize the loss \( \mathcal{L}(X, Z, A, Y) \) through a coordinated end-to-end learning process. Specifically, the optimal parameters for each module (\( \theta_D \), \( \theta_M \), \( \theta_R \)) are determined by minimizing the respective loss functions in a joint manner: first optimizing the neuro-symbolic learning component, followed by the neural architecture search component, and finally the disentangled representation learning component. This joint optimization framework enables the seamless integration of different modular components, achieving a more efficient and effective learning process that can be adapted to diverse tasks.\\
{\footnotesize Notations: $\mathcal{L}$: loss function (e.g., cross-entropy loss, MSE loss, etc.); $X$: input data; $Z$: disentangled representation; $A$: architecture of the modules; $Y$: target output, i.e., label; $\theta_D$, $\theta_M$, $\theta_R$: parameters of the representation, architecture and reasoning modules, respectively.}
}

\label{fig:framework}
\vspace{-1.5mm}
\end{figure*}

\section{A Unified Framework of MML for LLM}
\label{sec:mml-implementation}

\begin{figure}[t]
    \centering
        \includegraphics[width=0.4\textwidth]{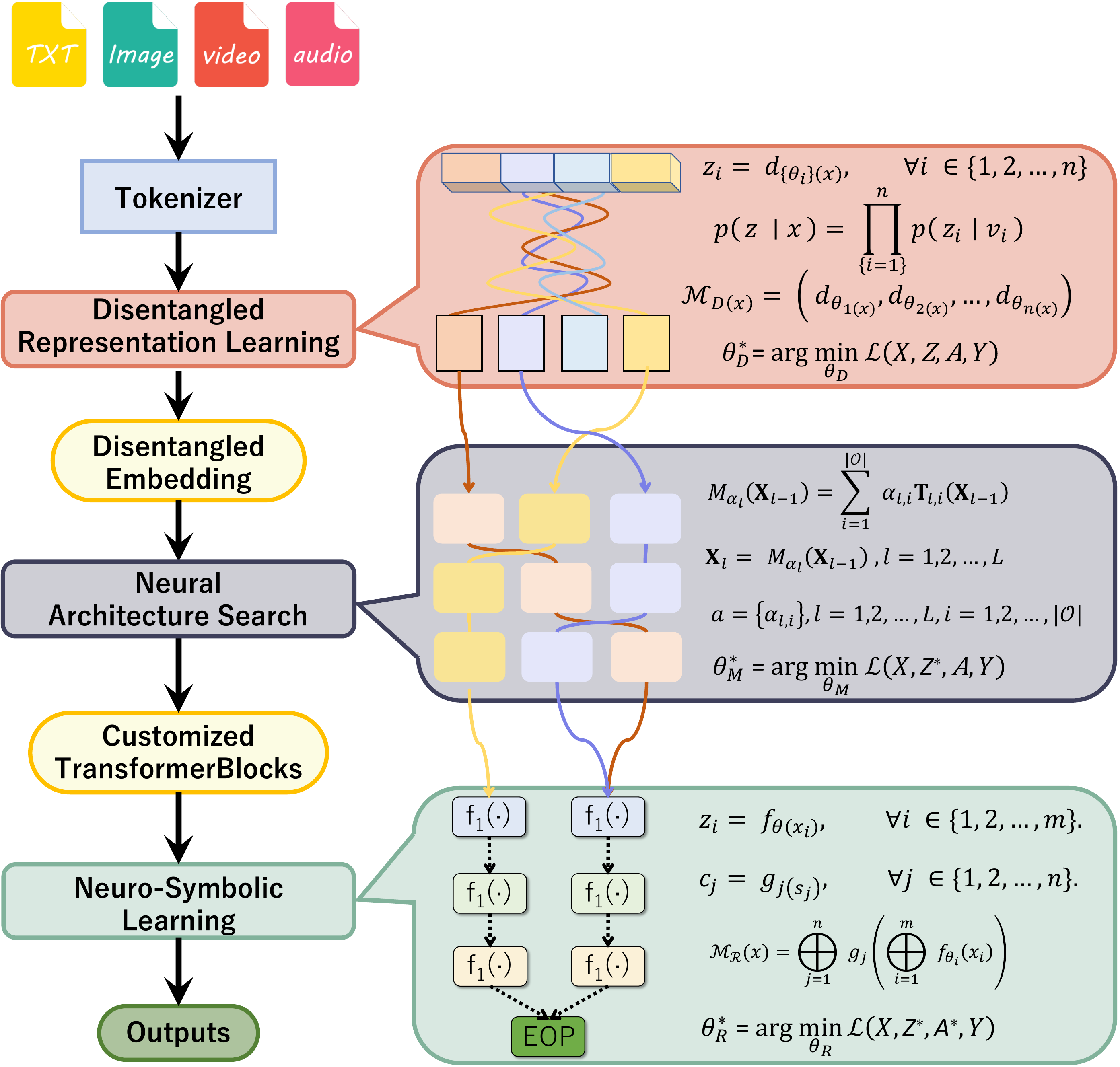}
    \caption{The overall learning pipeline of the proposed unified MML implementation: a joint optimization of: 1) Disentangled Representation Learning (DRL) module, which is responsible for learning a set of disentangled representations from the input data; 2) Neural Architecture Search (NAS) module, which is responsible for searching customized transformer blocks for different functions; and 3) Neuro-Symbolic Learning (NSL) module, which is responsible for refining the outputs of the NAS module through neuro-symbolic reasoning. 
    }
    \label{fig:method}
    \vspace{-3.5mm}
\end{figure}

Fig.~\ref{fig:framework} presents the general methodology of MML in detail and shows the unified framework with VQA as an example for instantiation. 
In this section, we continue to elaborate on the feasible implementation via introducing three complementary strategies: disentangled representation learning (DRL), neural architecture search (NAS), and neuro-symbolic learning (NSL).
First, DRL isolates independent semantic factors into modular representations, promoting transparency, generalizability, and controllability. The disentanglement enables the separation of complex data attributes, making the learned representations more interpretable and robust.
Second, NAS automates the design of modular networks by efficiently exploring large search spaces, and identifying optimal operator configurations and interconnections that strike a balance between performance and computational efficiency. This automation significantly reduces human effort in model design while enhancing flexibility.
Third, NSL integrates symbolic logic into the inference process within LLMs in a modular manner, enabling structured, rule-based validation and iterative refinement of neural outputs. This symbolic design mitigates issues such as hallucinations and reinforces logical consistency, becoming crucial for high-stakes applications.
Collectively, these interconnected strategies exemplify the practical implementation of MML. They not only address the inherent limitations of conventional LLMs but also pave the way for the next generation of robust, interpretable, and logic-capable models in various real-world applications.  
Fig.~\ref{fig:method} illustrates the overall optimization details of this MML implementation.

\subsection{Disentangled Representation Learning for Modular Representation}
\label{sec:disen}

Disentangled Representation Learning (DRL) for modular representation aims at separating and representing the underlying factors of variation in data independently and meaningfully. DRL seeks to break down complex data into distinct components or modules, each focusing on a specific aspect, such as color, shape, or size in images. Disentanglement is crucial in unsupervised and reinforcement learning as it facilitates effective decision-making through clear and meaningful representations. Unlike traditional end-to-end deep learning models that directly learn data representations capturing entangled features, DRL strives to extract latent variables that correspond to individual factors, fostering human-like generalization~\cite{geirhos2020shortcut}. 
The modular representation enables models to learn structured, explainable features, enhancing their generalization ability and adaptability to new situations~\cite{bengio2013representation}.

\noindent\textbf{Definition 1.} (Disentangled Representation Learning (DRL)) 
Let $\mathcal{M}_{\theta_D} = \{D_1, D_2, \dots, D_n\}$ represent a set of disentangled representation modules, where each module $D_i$ maps the input data $x \in \mathcal{X}$ to a latent variable $z_i$ through a function $d_{\theta_i}$, i.e.,  
$z_i = d_{\theta_i}(x), \quad \forall i \in \{1, 2, \dots, n\}.$

A \textit{disentangled embedding (a.k.a. representation)} $z = (z_1, z_2, \dots, z_n) \in \mathcal{Z}$ denotes the latent variables $z_i$ correspond to statistically independent factors of variation $v_i$ in the input data, such that:  
$p(z \mid x) = \prod_{i=1}^{n} p(z_i \mid v_i).$
This indicates that each latent variable $z_i$ is responsible for capturing only one underlying factor $v_i$ and remains invariant to changes in other factors $v_j$ ($j \neq i$). 
The modular representation function $\mathcal{M}_{\theta_D}$ can then be expressed as:  
\begin{align}\small\small
\mathcal{M}_{\theta_D}(x) = (d_{\theta_1}(x), d_{\theta_2}(x), \dots, d_{\theta_n}(x)).
\end{align}  
This function outputs a set of disentangled and semantically meaningful features.
The objective of DRL is to learn disentangled representations by minimizing the following loss function:  
\begin{align}\small\small
    & \theta_D^* = \arg\min_{\theta_D} \mathcal{L}(\mathbf{X}, \mathbf{Z}, \mathbf{A}, \mathbf{Y}), \\
    & \mathbf{Z}^* = \mathcal{M}_{\theta_D^*},
\end{align}
where $\mathbf{Z}$ is the disentangled representation and 
$\mathbf{Z}^*$ is the optimal learned representation, $\mathcal{L}$ denotes the loss function that measures the discrepancy between the predicted and the true outputs,
$\mathbf{X}$ is the input data, $\mathbf{A}$ is the architecture of the modules, and $\mathbf{Y}$ is the target output. The optimization process aims to learn the parameters $\theta_D$ of the disentangled representation modules such that the learned representations are semantically meaningful and independent.

This modular design highlights how DRL decomposes complex representations into independent, meaningful components, facilitating robust and interpretable learning. Specifically, DRL enables machine learning models to identify and disentangle hidden factors within observed data, thereby aligning learned representations with real-world semantics. These representations are invariant to external semantic changes~\cite{kim2018disentangling, lee2021learning}, aligned with real semantics, and robust against confounding or biased information~\cite{suter2019robustly}, making them suitable for diverse downstream tasks. 
DRL contributes to enhanced interpretability and efficiency in foundation models, such as ChatGPT and Stable Diffusion, by disentangling task-relevant knowledge from redundant components~\cite{zeng2022task}. This capability aids in making foundation models more transparent and adaptable, addressing challenges related to task specificity and interpretability. 

\subsection{Neural Architecture Search for Modular Model}
\label{sec:nas}

Neural Architecture Search (NAS) automates the design of neural network architectures, significantly reducing human effort while achieving state-of-the-art performance~\cite{elsken2019neural}. It is particularly valuable for modular model, where complex networks are broken down into specialized substructures or blocks that can be independently designed and optimized. NAS aims to identify the most effective combination of these modules and their interconnections, thus improving overall performance while maintaining computational efficiency. The modular model allows NAS to adapt to various tasks, such as image classification, object detection, semantic segmentation, and language processing, by tailoring network components to specific requirements~\cite{zoph2018learning, chen2019detnas, wang2020textnas}.

\noindent\textbf{Definition 2.} (Neural Architecture Search (NAS)) 
Let $\mathbf{M} = \{M_1, M_2, \dots, M_K\}$ represent a set of modular components, where each module $M_i$ corresponds to a specific function. These modules are selected through routing from a supernetwork. We define each layer in this supernetwork as follows:
\begin{align}\small
M_{\alpha_{l}}(\mathbf{X}_{l-1}) = \sum_{i=1}^{|\mathcal{O}|} \alpha_{l,i} \mathbf{T}_{l,i}(\mathbf{X}_{l-1}),
\end{align}
where $\mathbf{T}_{l,i}$ is the $i$-th block of the $l$-th layer, $\alpha_{l,i}$ is the weight of the $i$-th block of the $l$-th layer, and $\mathbf{X}_{l-1}$ is the input to the $l$-th layer, $|\mathcal{O}|$  is the number of blocks in the each layer. We can adopt a stack of layers to compose a module,
$\mathbf{X}_{l} = M_{\alpha_{l}}(\mathbf{X}_{l-1}), \quad l= 1,2,\dots,L,$
where  $L$ is the total number of layers.

Let the architecture search space $\mathcal{A}$ consist of all possible configurations formed by combining these blocks.
Let $a_j \in \mathcal{A} $ denote a specific architecture to form a module $M_{j}$ to achieve a specific function:
${a_j} = \{\alpha_{l,i}\}, \quad l= 1,2,\dots,L, i=1,2,\dots,|\mathcal{O}|,$
where $\alpha_{l,i}$ determines how the module $M_{j}$ weighs the blocks in each layer, and the group of them, i.e. $a_j$, is the configuration of the module $M_{j}$ that determines how it selects different calculation paths within the supernetwork to provide the flexibility of the model design.
 
The objective of NAS $\mathcal{M}_{\theta_M}$ is to find the different optimal architecture to form modular \textit{w.r.t.} the task. Let $\mathbf{A}$ denote the architecture set $\{ a_1, a_2, \dots, a_K \}$, where ${a_j}$ architecture for modular 
$M_j$. The optimization objective can be mathematically formulated as: 
\begin{align}\small
    &\theta_M^* = \arg\min_{\theta_M} \mathcal{L}(\mathbf{X}, \mathbf{Z}^*, \mathbf{A}, \mathbf{Y}), \\
    &\mathbf{A}^* = \mathcal{M}_{\theta_M^*},
\end{align}
where $\mathbf{A}$ is the chosen architecture and $\mathbf{A}^*$ is the optimal learned architecture, $\mathcal{L}$ denotes the loss function that measures the discrepancy between the predicted and the true outputs. The search process optimizes both the modules' design and their interconnections using search algorithms such as reinforcement learning, evolutionary algorithms, or gradient-based optimization. 
This modular design allows for the flexible combination of different neural components while efficiently identifying the optimal architecture, balancing between performance and computational cost. The NAS process typically comprises three key components: search space, search strategy, and evaluation strategy~\cite{elsken2019neural}. 

The search space in modular NAS is divided into macro and micro spaces. The macro space addresses the overall architecture, while the micro space focuses on individual modules or blocks. The modular design enables the integration of diverse operators such as convolution, attention, or aggregation, making it suitable for different data types like images, sequences, and graphs~\cite{zoph2016neural}.
The search strategies include reinforcement learning (RL), evolutionary algorithms (EA), and gradient-based methods. RL-based NAS treats the search process as a decision-making task, selecting modular components based on expected performance gains~\cite{jaafra2019reinforcement}. EA-based NAS evolves modular architectures by simulating biological evolution, iteratively selecting and mutating modules to enhance performance~\cite{de2016evolutionary}. Gradient-based NAS, such as DARTS~\cite{liu2018darts}, relaxes the search space into a continuous domain, allowing for efficient gradient optimization of modular connections and configurations.
Evaluation strategies aim to assess model performance efficiently, given the extensive search space. Techniques like weight sharing~\cite{xie2021weight}, predictor-based estimation~\cite{white2021powerful}, and zero-shot methods~\cite{chen2021bench} reduce the cost of evaluating different modular combinations. One-shot NAS, for instance, trains a super-network where all modular sub-networks share weights, enabling rapid evaluation of individual architectures without retraining~\cite{pham2018efficient}.

\subsection{Neuro-Symbolic Learning for Modular Reasoning}
\label{sec:nsr}

Neuro-Symbolic Learning (NSL) integrates the strengths of neural networks and symbolic reasoning to develop modular AI systems that are both flexible and interpretable~\cite{bengio2019system}. In modular reasoning, complex tasks are divided into smaller, specialized components, each responsible for a specific aspect of the problem. By leveraging neural networks' ability to learn from data and symbolic reasoning's structured logic, NSL enhances the strengths of both dynamic adaptability and logical reasoning, making it suitable for complex tasks in robotics, natural language processing, and autonomous systems~\cite{yi2018neural, dai2019bridging}. 

\noindent\textbf{Definition 3.} (Neuro-Symbolic Learning (NSL)) 

Let $\mathcal{M}_{\theta_R} = \{N_1, N_2, \dots, N_m\}$ represent a set of neural network modules, where each module $N_i$ maps input data $x_i$ to a latent representation $z_i$ through a function $f_{\theta_i}$, i.e.,
\begin{align}\small
z_i = f_{\theta_i}(x_i), \quad \forall i \in \{1, 2, \dots, m\}.
\end{align}
Let $\mathcal{S} = \{S_1, S_2, \dots, S_n\}$ represent a set of symbolic reasoning modules, where each symbolic module $S_j$ processes a symbolic representation $s_j$ to produce a logical conclusion $c_j$ through a reasoning function $g_j$, i.e., $c_j = g_j(s_j), \quad \forall j \in \{1, 2, \dots, n\}.$
The objective of NSL is to learn a joint modular model $\mathcal{M}$ that combines neural modules $\mathcal{N}$ and symbolic modules $\mathcal{S}$ as follows:
\begin{align}\small
\mathcal{M_R}(x) = \bigoplus_{j=1}^{n} g_j \left( \bigoplus_{i=1}^{m} f_{\theta_i}(x_i) \right).
\end{align}
Here, the symbol $\bigoplus$ denotes the modular aggregation operation, which combines the outputs from multiple modules into a unified representation. Specifically, the inner aggregation $\bigoplus_{i=1}^{m}$ collects the outputs from the neural modules, while the outer aggregation $\bigoplus_{j=1}^{n}$ combines the logical conclusions from the symbolic reasoning modules.
The model's optimization objective is to minimize the loss function:
\begin{align}\small
    &\theta_R^* = \arg\min_{\theta_R} \mathcal{L}(\mathbf{X}, \mathbf{Z}^*, \mathbf{A}, \mathbf{Y}), \\
    &\mathbf{Y}^* = \mathcal{M}_{\theta_R^*},
\end{align}
where $\mathbf{Y}$ is the output of the model, $\mathbf{X}$ is the input data, $\mathbf{A}$ is the architecture of the modules, and $\mathbf{Y}^*$ is the optimal learned output. The optimization process aims to learn the parameters $\theta_R$ of the neural and symbolic modules such that the combined model produces accurate predictions while maintaining interpretability and logical consistency.

This modular design fosters a balanced interaction between neural and symbolic systems. For instance, the ``System 1 and System 2'' framework by Yoshua Bengio~\footnote{\url{https://yoshuabengio.org/2023/03/21/scaling-in-the-service-of-reasoning-model-based-ml/}} outlines a collaborative model, where neural modules perform fast pattern recognition while symbolic modules conduct logical reasoning~\cite{bengio2019system}. The modular reasoning is able to optimize both neural networks and symbolic systems, enhancing the efficiency and interpretability of complex reasoning tasks~\cite{pfeiffer2023modular}. NSL also shows potential in improving LLMs by integrating logical consistency, guiding inference strategies, and enhancing prompt engineering, making LLMs more robust and interpretable~\cite{zhang2023siren}.

\section{Challenges and Future Directions}
\label{sec:future}

Although hybrid systems that integrate MML into LLMs promise enhanced reasoning, safety, and explainability, their development confronts several challenges illuminating pathways for future research.

\textbf{Integration and Configurable Interface.}
One of the foremost challenges is the seamless interface between neural networks---which operate in a continuous, gradient-driven space---and MML-based modules that execute discrete, logic-based operations. Current methods often depend on heuristic interfaces or differentiable relaxations to enable communication between these components. Moving forward, research should explore adaptive and dynamically configurable interfaces that can automatically modulate the degree of modular intervention based on input complexity or task demands. Advances in differentiable programming and co-training strategies may pave the way for truly unified architectures.

\textbf{Differentiability and Joint Optimization.}
The inherent mismatch between differentiable neural computations and non-differentiable modular reasoning creates significant obstacles for end-to-end training. Current solutions rely on surrogate loss functions or approximations, which can introduce biases and limit performance. Developing novel training methodologies such as meta-learning, reinforcement learning-guided structure search, or even fully differentiable modular reasoning paradigms can help to bridge this gap. Such approaches would enable more accurate backpropagation through modular components, ensuring that the combined system is able to learn coherently.

\textbf{Scalability and Computational Efficiency.}
LLMs are inherently computationally intensive, and the addition of modular reasoning layers may further escalate resource demands. This integration poses challenges for both training and real-time inference, particularly in resource-constrained environments or when processing high-volume data streams. Future research should prioritize the development of scalable architectures and efficient computational techniques for MML-based LLMs. Advances such as parameter-efficient fine-tuning, model quantization, and parallel processing frameworks can mitigate computational burdens while retaining the model performance.

\textbf{Evaluation and Benchmarking.}
Traditional evaluation metrics for LLMs, centered on language fluency and task-specific accuracy, fall short of capturing the logical consistency and interpretability gains afforded by MML. Without standardized benchmarks that assess both performance and reasoning quality, it is difficult to measure progress across different systems. Establishing comprehensive evaluation protocols that include metrics for logical coherence, safety, fairness, and explanation quality will be essential. Such benchmarks will not only facilitate rigorous comparisons but also drive the development of models that are robust in safety-critical and high-stakes applications.

\textbf{Balancing Interpretability and Performance.}
While integrating MML-based modules significantly enhances interpretability by providing clear audit trails and logical explanations, it often comes with trade-offs in raw predictive performance on data-intensive tasks. Striking a balance between maintaining high performance and achieving human-understandable reasoning remains an open research question. Future work might investigate adaptive systems that dynamically adjust the level of modular processing based on the confidence or complexity of the task, such as inference strategies that modulate reasoning depth in response to uncertainty measures.

\textbf{Modular World Models for Adaptive Reasoning.}
Another key future direction is the integration of Modular World Models (MWMs) to enhance reasoning, adaptability, and generalization with LLMs. Unlike traditional end-to-end deep learning approaches, MWMs decompose world knowledge into structured, reusable components that can dynamically interact with LLMs for causal inference, counterfactual reasoning, and real-world simulation. 
This modular approach enables efficient adaptation to new tasks, reducing the need for extensive retraining while improving interpretability and robustness. Future research should focus on a modular architecture that allows LLMs to query and update modular world models in the real-time world, leveraging advances in graph-based reasoning, differentiable programming, and structured knowledge distillation. By embedding MWMs, next-generation AI systems can move from static pattern recognition toward adaptive, compositional, and transparent decision-making in complex environments.

In summary, advancing MML-based LLMs requires improving integration mechanisms and training methodologies through adaptive interfaces, efficient joint optimization, scalable architectures, and standardized benchmarks, while balancing interpretability with performance. Such progress will be essential for building more robust, transparent, and versatile foundation models suited to real-world applications.

\section{Conclusions}

The rapid development of LLMs has revealed both their transformative potential and their limitations in reasoning, interpretability, and adaptability. This survey provides a systematic review of \textit{Modular Machine Learning (MML)}, organizing along two major dimensions: \textbf{Modular Data Representation} and \textbf{Modular Model Optimization}. We examine representative approaches under supervised and unsupervised settings, ranging from representation disentanglement to modular networks, symbolic constraints, neural architecture search, and inductive modularization, highlighting how modular principles enhance transparency, reliability, scalability, and extensibility compared to monolithic systems. Building on this taxonomy, we propose a unified framework of \textit{modular representation}, \textit{modular model}, and \textit{modular reasoning}, as well as discuss its significance for advancing next-generation LLMs. Looking ahead, open challenges include designing configurable interfaces, achieving efficient joint optimization, improving scalability, and establishing standardized benchmarks. In conclusion, we would like to argue that MML offers both a unifying perspective and a promising paradigm for guiding the evolution of reliable, interpretable, and adaptable large language models.

\footnotesize
\bibliographystyle{IEEEtran}
\bibliography{sn-bibliography}

{\scriptsize
\begin{IEEEbiography}[{\includegraphics[width=1in,height=1.25in,clip,keepaspectratio]{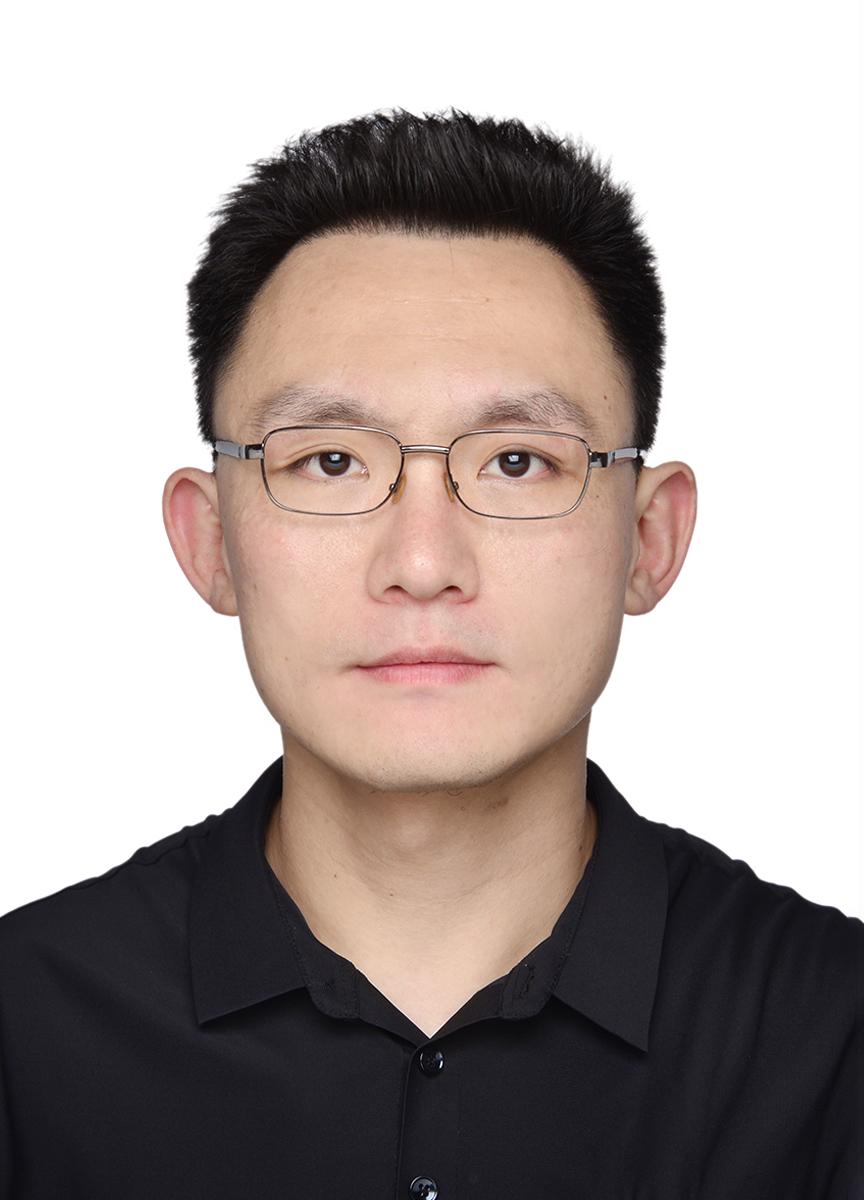}}]{Xin Wang} is currently an Associate Professor at the Department of Computer Science and Technology, Tsinghua University. He got both of his Ph.D. and B.E degrees in Computer Science and Technology from Zhejiang University, China. He also holds a Ph.D. degree in Computing Science from Simon Fraser University, Canada. His research interests include multimedia intelligence, machine learning and its applications. He has published over 200 high-quality research papers in ICML, NeurIPS, IEEE TPAMI, IEEE TKDE, ACM KDD, WWW, ACM SIGIR, ACM Multimedia etc., winning three best paper awards including ACM Multimedia Asia. He is the recipient of ACM China Rising Star Award, IEEE TCMC Rising Star Award and DAMO Academy Young Fellow.
\end{IEEEbiography}

\begin{IEEEbiography}[{\includegraphics[width=1in,height=1.25in,clip,keepaspectratio]{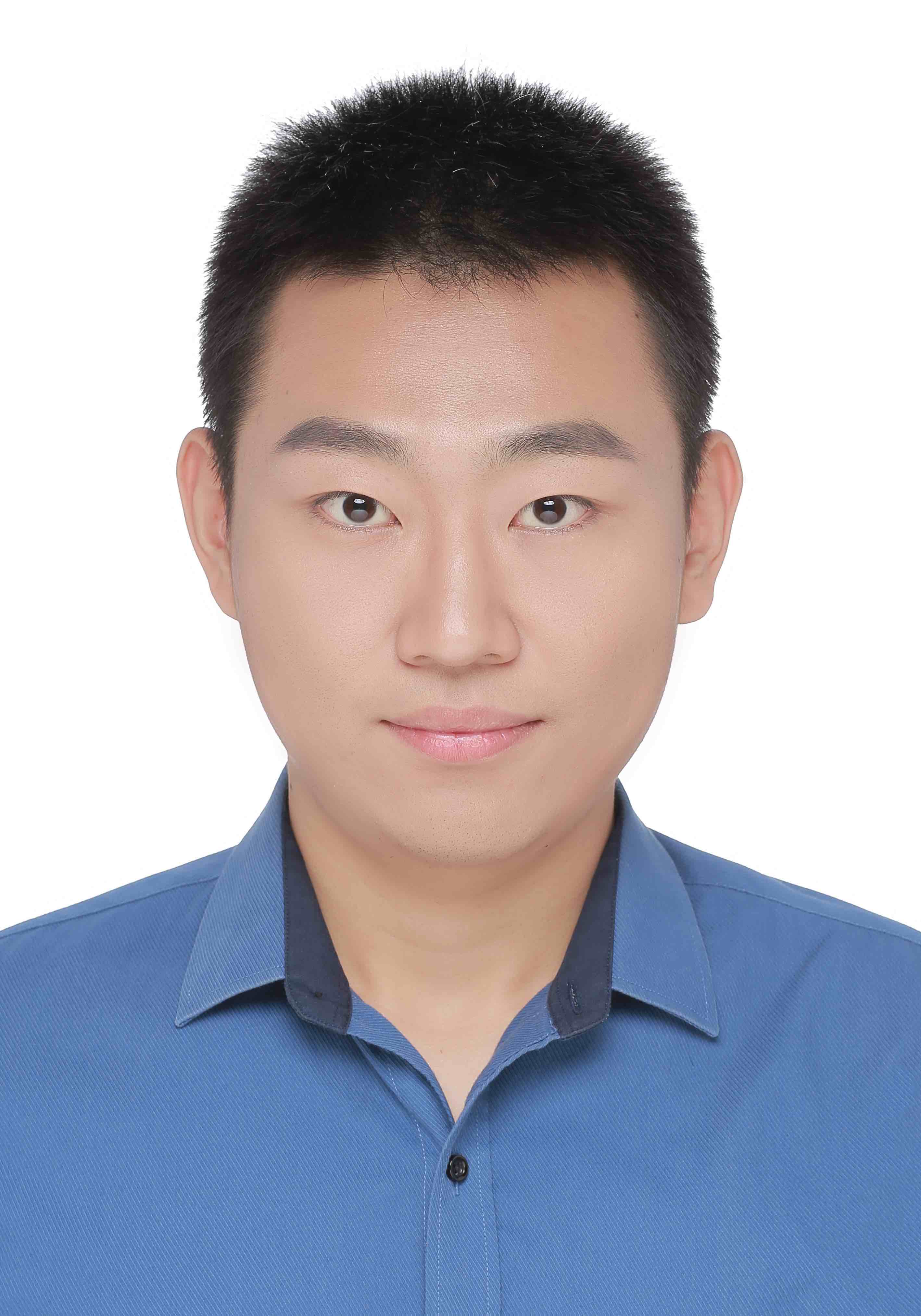}}]{Haoyang Li}  is currently a Postdoctoral Associate at Weill Cornell Medicine of Cornell University. He received his Ph.D. from the Department of Computer Science and Technology of Tsinghua University in 2023. He received his B.E. from the Department of Computer Science and Technology of Tsinghua University in 2018. His research interests are mainly in machine learning on graphs and out-of-distribution generalization. He has published high-quality papers in prestigious journals and conferences, e.g., IEEE TPAMI, TKDE, ACM TOIS, ICML, NeurIPS, ICLR, KDD, WWW, etc.
\end{IEEEbiography}

\begin{IEEEbiography}[{\includegraphics[width=1in,height=1.25in,clip,keepaspectratio]{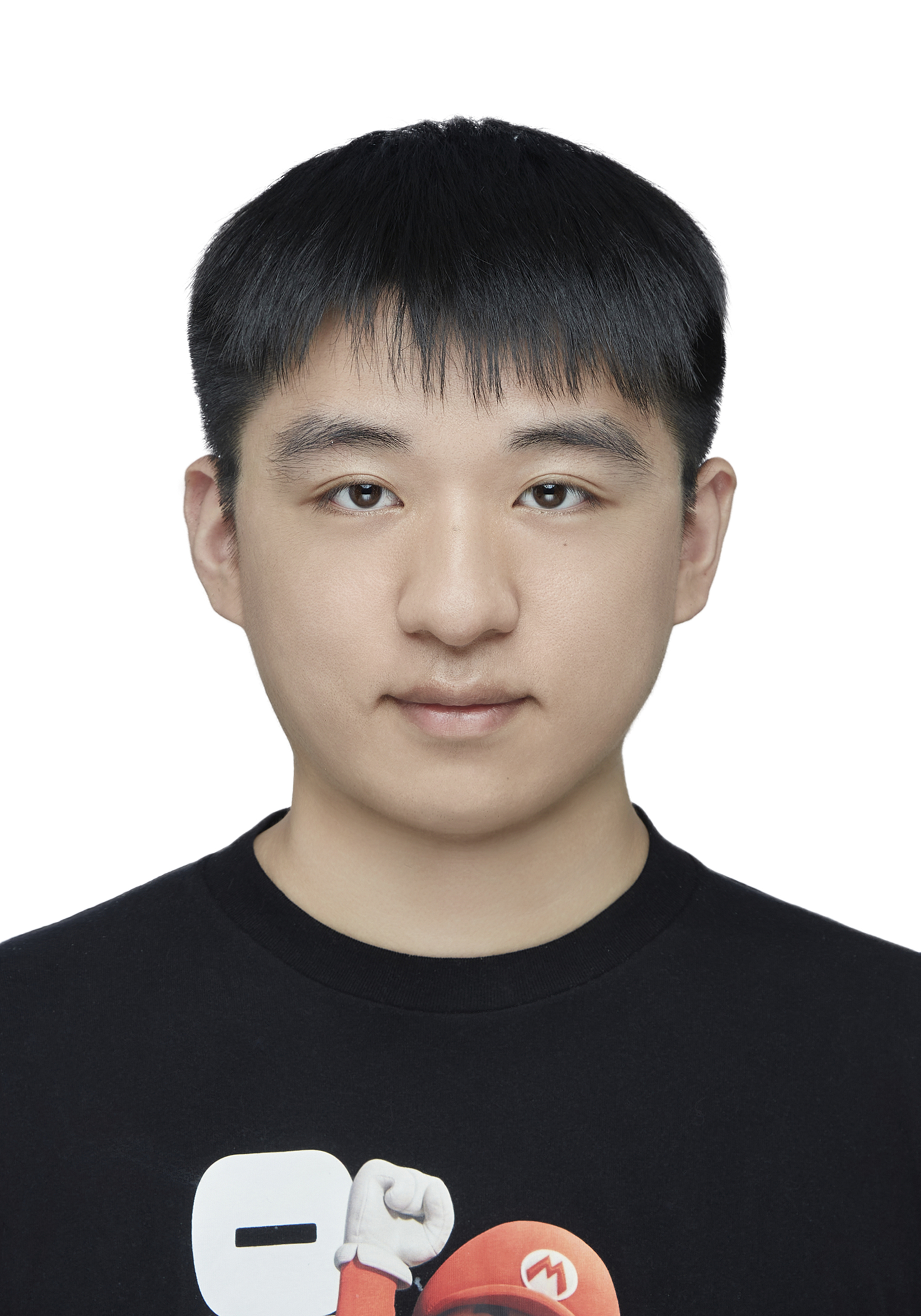}}]{Haibo Chen} is currently a Ph.D. student at the Department of Computer Science and Technology, Tsinghua University. He received his B.E. degree from the School of Computer Science and Engineering, Central South University. His main research interests include graph machine learning, out-of-distribution learning, and multi-modal graphs.
\end{IEEEbiography}

\begin{IEEEbiography}[{\includegraphics[width=1in,height=1.25in,clip,keepaspectratio]{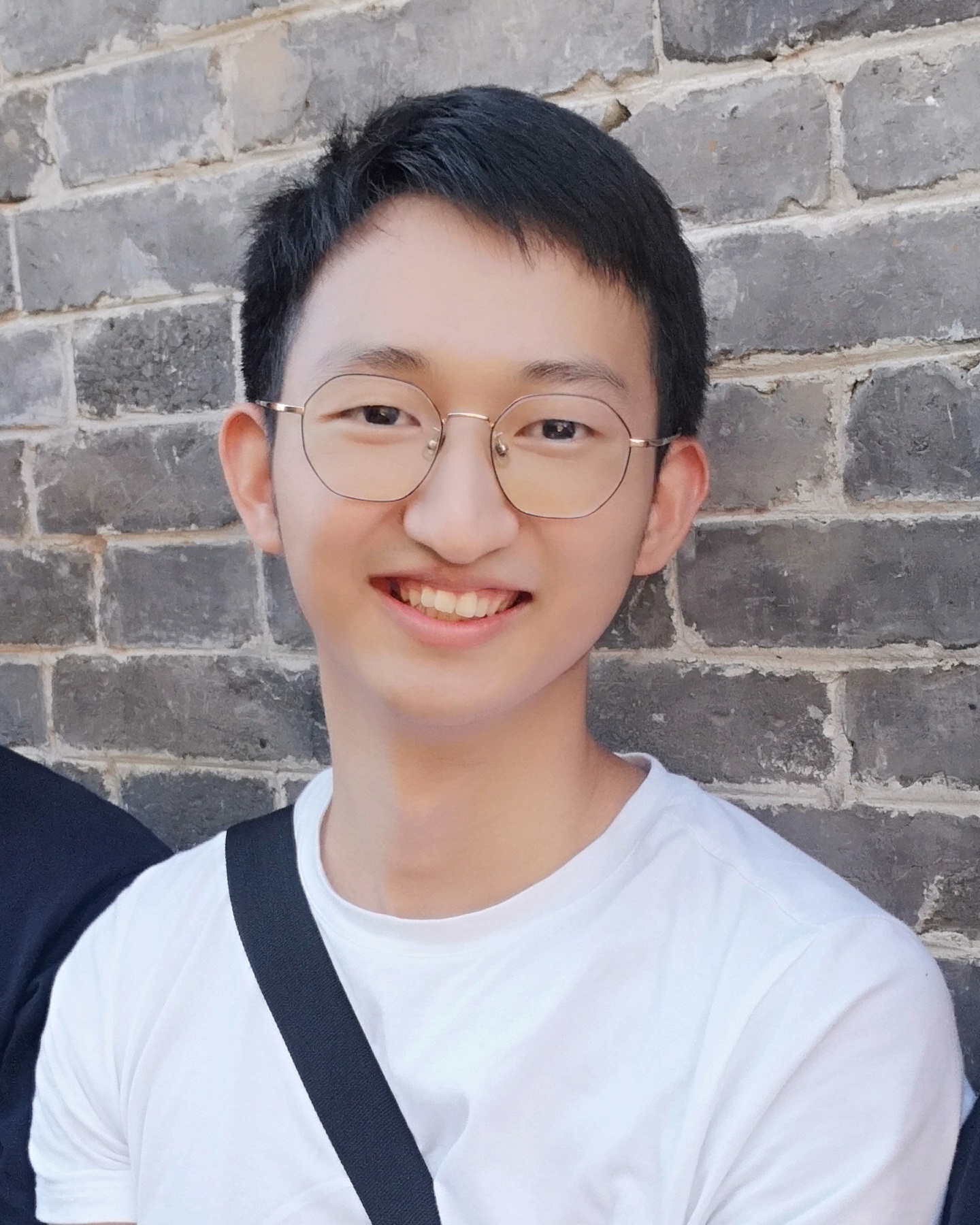}}]{Zeyang Zhang} received his B.E. from the Department of Computer Science and Technology, Tsinghua University in 2020. He is a Ph.D. candidate in the Department of Computer Science and Technology of Tsinghua University. His main research interests focus on graph representation learning, automated machine learning and out-of-distribution generalization. He has published several papers in prestigious conferences, e.g., NeurIPS, AAAI, etc.
\end{IEEEbiography}

\begin{IEEEbiography}[{\includegraphics[width=1in,height=1.25in,clip,keepaspectratio]{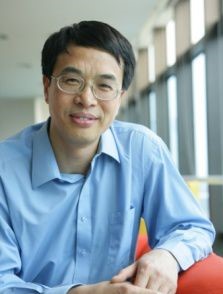}}]{Wenwu Zhu}
is currently a Professor in the Department of Computer Science and Technology at Tsinghua University. He also serves as the Vice Dean of National Research Center for Information Science and Technology, and the Vice Director of Tsinghua Center for Big Data. Prior to his current post, he was a Senior Researcher and Research Manager at Microsoft Research Asia. He was the Chief Scientist and Director at Intel Research China from 2004 to 2008. He worked at Bell Labs, New Jersey as Member of Technical Staff during 1996-1999. He received his Ph.D. degree from New York University in 1996. 
His research interests are in the area of data-driven multimedia networking and Cross-media big data computing. He has published over 400 referred papers and is the inventor or co-inventor of over 100 patents. He received eight Best Paper Awards, including ACM Multimedia 2012 and IEEE Transactions on Circuits and Systems for Video Technology in 2001 and 2019.  

He served as EiC for IEEE Transactions on Multimedia from 2017-2019. He served in the steering committee for IEEE Transactions on Multimedia (2015-2016) and IEEE Transactions on Mobile Computing (2007-2010), respectively. He serves as General Co-Chair for ACM Multimedia 2018 and ACM CIKM 2019, respectively. 
He is an AAAS Fellow, IEEE Fellow, SPIE Fellow, and a member of The Academy of Europe (Academia Europaea).
\end{IEEEbiography}
}
\end{document}